\PassOptionsToPackage{dvipsnames}{xcolor}

\documentclass{article} %
\usepackage{iclr2024_conference,times}

\usepackage{amsmath,amsfonts,bm}

\newcommand{\figleft}{{\em (Left)}}

\newcommand{\figright}{{\em (Right)}}
\newcommand{\figtop}{{\em (Top)}}
\newcommand{\figbottom}{{\em (Bottom)}}

\def\eqref#1{equation~\ref{#1}}

\def\1{\bm{1}}

\DeclareMathAlphabet{\mathsfit}{\encodingdefault}{\sfdefault}{m}{sl}
\SetMathAlphabet{\mathsfit}{bold}{\encodingdefault}{\sfdefault}{bx}{n}

\def\gA{{\mathcal{A}}}

\def\gL{{\mathcal{L}}}

\def\gS{{\mathcal{S}}}
\def\gT{{\mathcal{T}}}

\def\gX{{\mathcal{X}}}
\def\gY{{\mathcal{Y}}}

\newcommand{\E}{\mathbb{E}}

\newcommand{\R}{\mathbb{R}}

\DeclareMathOperator*{\argmax}{arg\,max}
\DeclareMathOperator*{\argmin}{arg\,min}

\usepackage[colorlinks]{hyperref}
\usepackage{url}
\usepackage{cancel}
\usepackage{graphicx}
\usepackage{amssymb}
\usepackage{bbm}
\usepackage{amsthm}
\usepackage{algorithm}
\usepackage[noend]{algpseudocode}
\usepackage{caption}
\usepackage{subcaption}
\usepackage{wrapfig}
\usepackage{booktabs}
\usepackage{mathtools}
\usepackage{xcolor}
\usepackage{pifont}

\newcommand{\heavyxmark}{\ding{54}}%
\newcommand{\starmark}{\ding{72}}%
\newcommand{\CE}{\mathcal{CE}}

\definecolor{mypurple}{HTML}{7A4988}
\definecolor{mygreen}{HTML}{2ca02c}
\definecolor{myblue}{HTML}{1f77b4}
\hypersetup{urlcolor=myblue, citecolor=myblue, linkcolor=myblue}

\DeclarePairedDelimiter\abs{\lvert}{\rvert}%

\title{Contrastive Difference Predictive Coding}

\author{Chongyi Zheng \\
Carnegie Mellon University \\
\texttt{chongyiz@andrew.cmu.edu} \\
\And
Ruslan Salakhutdinov \\
Carnegie Mellon University \\
\And
Benjamin Eysenbach \\
Princeton University \\
}

\iclrfinalcopy %
\begin{document}

\maketitle

\begin{abstract}

Predicting and reasoning about the future lie at the heart of many time-series questions. For example, goal-conditioned reinforcement learning can be viewed as learning representations to predict which states are likely to be visited in the future. While prior methods have used contrastive predictive coding to model time series data, learning representations that encode long-term dependencies usually requires large amounts of data. In this paper, we introduce a temporal difference version of contrastive predictive coding that stitches together pieces of different time series data to decrease the amount of data required to learn predictions of future events. We apply this representation learning method to derive an off-policy algorithm for goal-conditioned RL. Experiments demonstrate that, compared with prior RL methods, ours achieves $2 \times$ median improvement in success rates and can better cope with stochastic environments. In tabular settings, we show that our method is about $20\times$ more sample efficient than the successor representation and $1500 \times$ more sample efficient than the standard (Monte Carlo) version of contrastive predictive coding.

\textbf{Code}: {\footnotesize\url{https://github.com/chongyi-zheng/td_infonce}}\\
\textbf{Website}: {\footnotesize\url{https://chongyi-zheng.github.io/td_infonce}}

\end{abstract}

\section{Introduction}
\label{sec:intro}

\begin{wrapfigure}{R}{0.5\textwidth}
    \centering
    \includegraphics[width=\linewidth]{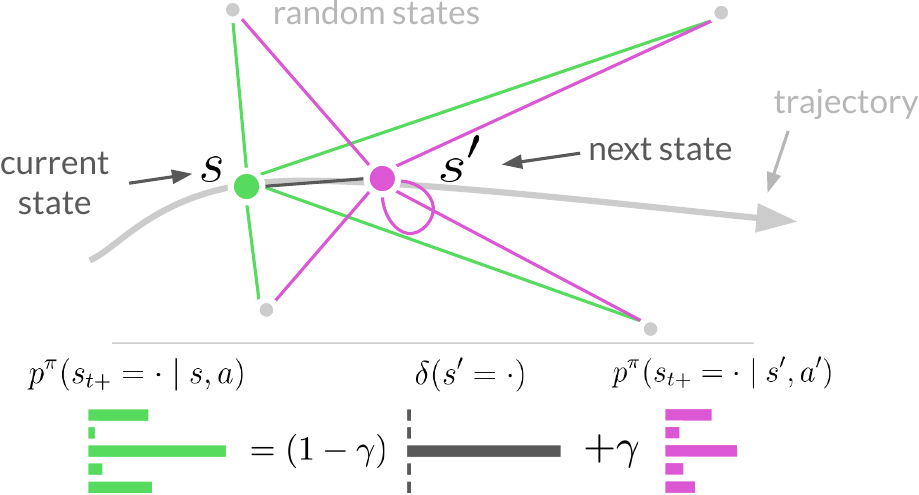}
    \caption{\footnotesize \textbf{TD InfoNCE} is a nonparametric version of the successor representation. 
    \figtop \, The distances between learned representations indicate the probability of transitioning to the next state and a set of randomly-sampled states. 
    \figbottom \, We update these representations so they assign high likelihood to \emph{(a)} the next state and \emph{(b)} states likely to be visited after the next state. See Sec.~\ref{sec:method} for details.}
    \label{fig:method}
    \vspace{-1em}
\end{wrapfigure}

Learning representations is important for modeling high-dimensional time series data. Many applications of time-series modeling require representations that not only contain information about the contents of a particular observation, but also about how one observation relates to others that co-occur in time.
Acquiring representations that encode temporal information is challenging, especially when attempting to capture long-term temporal dynamics: the frequency of long-term events may decrease with the time scale, meaning that learning longer-horizon dependencies requires larger quantities of data.

In this paper, we study contrastive representation learning on time series data -- positive examples co-occur nearby in time, so the distances between learned representations should encode the likelihood of transiting from one representation to another.
Building on prior work that uses the InfoNCE~\citep{sohn2016improved, oord2018representation} loss to learn representations of time-series data effectively, we will aim to build a temporal difference version of this loss.
Doing so allows us to optimize this objective with fewer samples, enables us to stitch together pieces of different time series data, and enables us to perform counterfactual reasoning -- we should be able to estimate which representations we would have learned, if we had collected data in a different way. After a careful derivation, our resulting method can be interpreted as a non-parametric form of the successor representation~\citep{dayan1993improving}, as shown in Fig.~\ref{fig:method}.

The main contribution of this paper is a temporal difference estimator for InfoNCE. We then apply this estimator to develop a new algorithm for goal-conditioned RL. Experiments on both state-based and image-based benchmarks show that our algorithm outperforms prior methods, especially on the most challenging tasks. Additional experiments demonstrate that our method can handle stochasticity in the environment more effectively than prior methods. We also demonstrate that our algorithm can be effectively applied in the offline setting. Additional tabular experiments demonstrate that TD InfoNCE is up to $1500 \times$ more sample efficient than the standard Monte Carlo version of the loss and that it can effectively stitch together pieces of data.

\section{Related Work}
\label{sec:prior-work}

This paper will study the problem of self-supervised RL, building upon prior methods on goal-condition RL, contrastive representation learning, and methods for predicting future state visitations. Our analysis will draw a connection between these prior methods, a connection which will ultimately result in a new algorithm for goal-conditioned RL. We discuss connections with unsupervised skill learning and mutual information in Appendix~\ref{appendix:mi}.

\paragraph{Goal-conditioned reinforcement learning.} Prior work has proposed many frameworks for learning goal-conditioned policies, including conditional supervised learning~\citep{ding2019goal, ghosh2020learning, gupta2020relay, emmons2021rvs, lynch2020learning, oh2018self, sun2019policy}, actor-critic methods~\cite{andrychowicz2017hindsight, nachum2018data, chane2021goal}, semi-parametric planning~\citep{pertsch2020long, fang2022planning, fang2023generalization, eysenbach2019search, nair2019hierarchical, gupta2020relay}, and distance metric learning~\citep{pmlr-v202-wang23al, tian2020model, nair2020goal, durugkar2021adversarial}.
These methods have demonstrated impressive results on a range of tasks, including real-world robotic tasks~\citep{ma2022vip, shah2022rapid, zheng2023stabilizing}.
While some methods require manually-specified reward functions or distance functions, our work builds upon a self-supervised interpretation of goal-conditioned RL that casts this problem as predicting which states are likely to be visited in the future~\citep{eysenbach2020c, eysenbach2022contrastive, blier2021learning}.%

\paragraph{Contrastive representation learning.} Contrastive learning methods have become a key tool for learning representations in computer vision and NLP~\citep{chopra2005learning, schroff2015facenet, sohn2016improved, oh2016deep, wang2020understanding, oord2018representation, tschannen2019mutual, weinberger2009distance, he2022masked, radford2021learning, chen2020simple, tian2020contrastive, gao2021simcse}. These methods assign similar representations to positive examples and dissimilar representations to negative examples or outdated embeddings~\citep{grill2020bootstrap}.
The two main contrastive losses are based on binary classification (``NCE'') ranking loss (``InfoNCE'')~\citep{ma2018noise}. Modern contrastive learning methods typically employ the ranking-based objective to learn representations of images~\citep{chen2020simple, tian2020contrastive, henaff2020data, wu2018unsupervised}, text~\citep{logeswaran2018efficient, jia2021scaling, radford2021learning} and sequential data~\citep{nair2022r3m, sermanet2018time}.
Prior works have also provided theoretical analysis for these methods from the perspective of mutual information maximization~\citep{linsker1988self, poole2019variational}, noise contrastive estimation~\citep{gutmann2010noise, ma2018noise, tsai2020neural, arora2019theoretical}, and the geometry of the learned representations~\citep{wang2020understanding}. 
In the realm of RL, prior works have demonstrated that contrastive methods can provide effective reward functions and auxiliary learning objectives~\citep{laskin2020curl, laskin2020reinforcement, hansen2022bisimulation, choi2021variational, nair2020contextual, nair2018visual}, and can also be used to formulate the goal-reaching problem in an entirely self-supervised manner~\citep{ma2022vip, durugkar2021adversarial, eysenbach2020c, eysenbach2022contrastive}. 
Our method will extend these results by building a temporal difference version of the ``ranking''-based contrastive loss; this loss will enable us to use data from one policy to estimate which states a different policy will visit.

\paragraph{Temporal difference learning and successor representation.} Another line of work studies using temporal difference learning to predict states visited in the future, building upon successor representations and successor features~\citep{dayan1993improving, barreto2017successor, barreto2019option, blier2021learning}. While learning successor representation using temporal difference bears a similarity to the typical Q-Learning algorithm~\citep{watkins1992q, fu2019diagnosing, mnih2015human} in the tabular setting,  directly estimating this quantity is difficult with continuous states and actions~\citep{janner2020gamma, barreto2017successor, touati2021learning, blier2021learning}. To lift this limitation, we will follow prior work~\citep{eysenbach2022contrastive, eysenbach2020c, touati2021learning} in predicting the successor representation indirectly: rather than learning a representation whose coordinates correspond to visitation probabilities, we will learn state representations such that their inner product corresponds to a visitation probability. Unlike prior methods, we will show how the common InfoNCE objective can be estimated in a temporal difference fashion, opening the door to off-policy reasoning and enabling our method to reuse historical data to improve data efficiency.

\section{Method}
\label{sec:method}

We start by introducing notation and prior approaches to the contrastive representation learning and the goal-conditioned RL problems. We then propose a new self-supervised actor-critic algorithm that we will use in our analysis.

\subsection{Preliminaries}
We first review prior work in contrastive representation learning and goal-conditioned RL. Our method will use ideas from both.

\paragraph{Contrastive representation via InfoNCE.}
Contrastive representation learning aims to learn a representation space, pushing representations of positive examples together and pushing representations of negative examples away. InfoNCE (also known as contrastive predictive coding)~\citep{sohn2016improved, jozefowicz2016exploring, oord2018representation, henaff2020data} is a widely used contrastive loss, which builds upon noise contrastive estimation (NCE)~\citep{gutmann2010noise, ma2018noise}. 
Given the distribution of data $p_{\gX}(x), p_{\gY}(y)$ over data $x \in \gX, y \in \gY$ and the conditional distribution of positive pairs $p_{\gY \mid \gX}(y | x)$ over $\gX \times \gY$, we sample $x \sim p_{\gX}(x)$, $y^{(1)} \sim p_{\gY \mid \gX}(y \mid x)$, and $\{y^{(2)}, \cdots, y^{(N)}\} \sim p_{\gY}(y)$. The InfoNCE loss is defined as
\begin{align}
    \mathcal{L}_{\text{InfoNCE}}(f) \triangleq  \mathbb{E}_{\substack{x \sim p_{\gX}(x), y^{(1)} \sim p_{\gY \mid \gX}(y \mid x) \\ y^{(2:N)} \sim p_{\gY}(y)}} \left[ \log \frac{e^{f(x, y^{(1)})}}{ \sum_{i = 1}^N e^{f(x, y^{(i)})} } \right],
    \label{eq:infonce}
\end{align}
where $f: \gX \times \gY \mapsto \R$ is a parametric function. Following prior work~\citep{eysenbach2022contrastive, wang2020understanding, touati2021learning}, we choose to parameterize $f(\cdot, \cdot)$ via the inner product of representations of data $f(x, y) = \phi(x)^{\top} \psi(y)$, where $\phi(\cdot)$ and $\psi(\cdot)$ map data to $\ell_2$ normalized vectors of dimension $d$. We will call $f$ the \emph{critic function} and $\phi$ and $\psi$ the \emph{contrastive representations}. The Bayes-optimal critic for the InfoNCE loss satisfies~\citep{poole2019variational, ma2018noise, oord2018representation}
\begin{align*}
    \exp \left(f^{\star}(x, y) \right) = \frac{p(y \mid x)}{p(y) c(x)},
\end{align*}
where $c(\cdot)$ is an arbitrary function. We can estimate this arbitrary function using the optimal critic $f^{\star}$ by sampling multiple negative pairs from the data distribution:
\begin{align}
    \E_{p(y)}\left[\exp \left( f^{\star}(x, y) \right) \right] = \int \cancel{p(y)} \frac{p(y \mid x)}{\cancel{p(y)}c(x)} dy = \frac{1}{c(x)} \underbrace{\int p(y \mid x) dy}_{ = 1} = \frac{1}{c(x)}.
    \label{eq:arbitrary-func-est}
\end{align}

\paragraph{Reinforcement learning and goal-conditioned RL.} We will consider a Markov decision process defined by states $s \in \gS$, actions $a \in \gA$, rewards $r: \gS \times \gA \times \gS \mapsto \R$. %
Using  $\Delta(\cdot)$ denotes the probability simplex, we define an initial state distribution $p_0: \mathcal{S} \mapsto \Delta(\mathcal{S})$,  discount factor $\gamma \in (0, 1]$, and  dynamics $p: \mathcal{S} \times \mathcal{A} \mapsto \Delta(\mathcal{S})$.
Given a policy $\pi: \gS \mapsto \Delta(\gA)$, we will use $p^{\pi}_t(s_{t+} \mid s, a)$ to denote the probability density of reaching state $s_{t+}$ after exactly $t$ steps, starting at state $s$ and action $a$ and then following the policy $\pi(a \mid s)$. We can then define the discounted state occupancy measure~\citep{ho2016generative, zhang2020gradientdice, eysenbach2020c, eysenbach2022contrastive, zheng2023stabilizing} starting from state $s$ and action $a$ as 
\begin{align}
    p^{\pi}(s_{t+} \mid s, a) \triangleq (1 - \gamma) \sum_{t = 1}^{\infty} \gamma^{t - 1} p_t^{\pi}(s_{t+} \mid s, a).
    \label{eq:discounted-state-occupancy-measure}
\end{align}
Prior work~\citep{dayan1993improving} have shown that this discounted state occupancy measure follows a recursive relationship between the density at the current time step and the future time steps:
\begin{align}
    p^{\pi}(s_{t+} \mid s, a) = (1 - \gamma) p(s' = s_{t+} \mid s, a) + \gamma \mathbb{E}_{\substack{s' \sim p(s' \mid s, a) \\ a' \sim \pi(a' \mid s')}} \left[ p^{\pi}(s_{t+} \mid s', a') \right].
    \label{eq:discounted-state-occupancy-measure-recurrence}
\end{align}
For goal-conditioned RL, we define goals $g \in \gS$ in the same space as states and consider a  goal-conditioned policy $\pi(a \mid s, g)$ and the corresponding goal-conditioned discounted state occupancy measure $p^{\pi}(s_{t+} \mid s, a, g)$. For evaluation, we will sample goals from a distribution $p_g: \gS \mapsto \Delta(\gS)$. Following prior work~\citep{eysenbach2020c, rudner2021outcome}, we define the objective of the goal-reaching policy as maximizing the probability of reaching desired goals under its discounted state occupancy measure while commanding the same goals:
\begin{align}
    \max_{\pi(\cdot \mid \cdot, \cdot)} \mathbb{E}_{p_g(g),  p_0(s), \pi(a \mid s, g)} \left[ p^{\pi}(s_{t+} = g \mid s, a, g) \right].
    \label{eq:policy-obj}
\end{align}
In tabular settings, this objective is the same as maximizing expected returns using a sparse reward function $r(s, a, s', g) = (1 - \gamma) \delta(s' = g)$~\citep{eysenbach2022contrastive}.
Below, we review two strategies for estimating the discounted state occupancy measure. Our proposed method (Sec.~\ref{subsec:td-infonce}) will combine the strengths of these methods while lifting their respective limitations.

\paragraph{Contrastive RL and C-Learning.}
Our focus will be on using contrastive representation learning to build a new goal-conditioned RL algorithm, following a template set in prior work~\citep{eysenbach2022contrastive, eysenbach2020c}.
These \emph{contrastive RL} methods are closely related to the successor representation~\citep{dayan1993improving}: they aim to learn representations whose inner products correspond to the likelihoods of reaching future states.
Like the successor representation, representations from these contrastive RL methods can then be used to represent the Q function for any reward function~\citep{mazoure2022contrastive}.
Prior work~\citep{eysenbach2022contrastive} has shown how both NCE and the InfoNCE losses can be used to derive Monte Carlo algorithms for estimating the discounted state occupancy measure. We review the Monte Carlo InfoNCE loss below. Given a policy $\pi(a \mid s)$, consider learning contrastive representations for a state and action pair $x = (s, a)$ and a potential future state $y = s_{t+}$. We define the data distribution to be the joint distribution of state-action pairs $p_{\gX}(x) = p(s, a)$ and the marginal distribution of future states $p_{\gY}(y) = p(s_{t+})$, representing either the distribution of a replay buffer (online) or the distribution of a dataset (offline). The conditional distribution of positive pairs is set to the discounted state occupancy measure for policy $\pi$, $p_{\gY \mid \gX}(y \mid x) = p^{\pi}(s_{t+} \mid s, a)$, resulting in a Monte Carlo (MC) estimator
\begin{align}
    \mathcal{L}_{\text{MC InfoNCE}}(f) = \mathbb{E}_{\substack{(s, a) \sim p(s, a), s_{t+}^{(1)} \sim p^{\pi}(s_{t+} \mid s, a) \\ s_{t+}^{(2:N)} \sim p(s_{t+})}} \left[ \log \frac{e^{f(s, a, s_{t+}^{(1)})}}{ \sum_{i = 1}^N e^{f(s, a, s_{t+}^{(i)})} } \right]
    \label{eq:mc-infonce}
\end{align}
and an optimal critic function satisfying
\begin{align}
    \exp(f^{\star}(s, a, s_{t+})) = \frac{p^{\pi}(s_{t+} \mid s, a)}{p(s_{t+}) c(s, a)}.
    \label{eq:opt-critic}
\end{align}
This loss estimates the discounted state occupancy measure in a Monte Carlo manner. Computing this estimator usually requires sampling future states from the discounted state occupancy measure of the policy $\pi$, i.e., on-policy data. While, in theory, Monte Carlo estimator can be used in an off-policy manner by applying importance weights to correct actions, this estimator usually suffers from high variance and is potentially sample inefficient than temporal difference methods~\citep{precup2000eligibility, precup2001off}.

In the same way that temporal difference (TD) algorithms tend to be more sample efficient than Monte Carlo algorithms for reward maximization~\citep{sutton2018reinforcement}, we expect that TD contrastive methods are more sample efficient at estimating probability ratios than their Monte Carlo counterparts.
Given that the InfoNCE tends to outperform the NCE objective in other machine learning disciplines, we conjecture that our TD InfoNCE objective will outperform the TD NCE objective~\citep{eysenbach2020c} (see experiments in Appendix.~\ref{appendix:td-infonce-vs-c-learning}).

\subsection{Temporal Difference InfoNCE}
\label{subsec:td-infonce}

In this section, we derive a new loss for estimating the discounted state occupancy measure for a fixed policy. This loss will be a temporal difference variant of the InfoNCE loss. We will use \textbf{temporal difference InfoNCE (TD InfoNCE)} to refer to our loss function. %

In the off-policy setting, we aim to estimate the discounted state occupancy measure of the policy $\pi$ given a dataset of transitions $\mathcal{D} = \left\{(s, a, s')_{i} \right\}_{i = 1}^{D}$ collected by another behavioral policy $\beta(a \mid s)$. This setting is challenging because we do not obtain samples from the discounted state occupancy measure of the target policy $\pi$. Addressing this challenge involves two steps:~\emph{(i)} expanding the MC estimator (Eq.~\ref{eq:mc-infonce}) via the recursive relationship of the discounted state occupancy measure (Eq.~\ref{eq:discounted-state-occupancy-measure-recurrence}), and~\emph{(ii)} estimating the expectation over the discounted state occupancy measure via importance sampling. We first use the identity from Eq.~\ref{eq:discounted-state-occupancy-measure-recurrence} to express the MC InfoNCE loss as the sum of a next-state term and a future-state term:
\begin{align*}
     &\mathbb{E}_{\substack{ (s, a) \sim p(s, a) \\ s_{t+}^{(2:N)} \sim p(s_{t+}) }} \Bigg[ (1 - \gamma) \underbrace{ \mathbb{E}_{s_{t+}^{(1)} \sim p(s' \mid s, a)} \left[ \log \frac{ e^{ f(s, a, s_{t+}^{(1)}) } }{\sum_{i = 1}^N e^{ f(s, a, s_{t+}^{(i)}) }} \right]}_{ \mathcal{L}_1(f) }  \nonumber \\
    & \hspace{6.5em} + \gamma \underbrace{ \mathbb{E}_{ \substack{s' \sim p(s' \mid s, a), a' \sim \pi(a' \mid s') \\ s_{t+}^{(1)} \sim p^{\pi}(s_{t+} \mid s', a')}} \left[ \log \frac{ e^{ f(s, a, s_{t+}^{(1)}) } }{\sum_{i = 1}^N e^{ f(s, a, s_{t+}^{(i)}) }} \right] }_{\mathcal{L}_2(f)} \Bigg].
\end{align*}
While this estimate is similar to a TD target for Q-Learning~\citep{watkins1992q, fu2019diagnosing}, the second term requires sampling from the discounted state occupancy measure of policy $\pi$. To avoid this sampling, we next replace the expectation over $p^{\pi}(s_{t+} \mid s', a')$ in $\mathcal{L}_2(f)$ by an importance weight, 
\begin{align*}
    \gL_2(f) &= \mathbb{E}_{ \substack{s' \sim p(s' \mid s, a), a' \sim \pi(a' \mid s') \\ s_{t+}^{(1)} \sim p(s_{t+})}} \left[ \frac{ p^{\pi}(s_{t+}^{(1)} \mid s', a') }{p(s_{t+}^{(1)})} \log \frac{ e^{ f(s, a, s_{t+}^{(1)}) } }{\sum_{i = 1}^N e^{ f(s, a, s_{t+}^{(i)}) }} \right].
\end{align*}
If we could estimate the importance weight, then we could easily estimate this term by sampling from $p(s_{t+})$. We will estimate this importance weight by rearranging the expression for the optimal critic (Eq.~\ref{eq:opt-critic}) and substituting our estimate for the normalizing constant $c(s, a)$ (Eq.~\ref{eq:arbitrary-func-est}):
\begin{align}
    \frac{ p^{\pi}(s_{t+}^{(1)} \mid s, a) }{p(s_{t+}^{(1)})} &=  c(s, a) \cdot \exp \left(f^\star(s, a, s_{t+}^{(1)}) \right) = \frac{e^{f^{\star}(s, a, s_{t+}^{(1)})} }{\E_{p(s_{t+})}\left[e^{f^{\star}(s, a, s_{t+})}\right]}.
\end{align}
We will use $w(s, a, s_{t+}^{(1:N)})$ to denote our estimate of this, using $f$ in place of $f^\star$ and using a finite-sample estimate of the expectation in the denominator:
\begin{align}
    w(s, a, s_{t+}^{(1:N)}) \triangleq \frac{e^{f(s, a, s_{t+}^{(1)}) }}{ \frac{1}{N}\sum_{i = 1}^N e^{ f(s, a, s_{t+}^{(i)}) } }
    \label{eq:importance-weight}
\end{align}
This weight accounts for the effect of the discounted state occupancy measure of the target policy. Additionally, it corresponds to the categorical classifier that InfoNCE produces (without constant $N$). Taken together, we can now substitute the importance weight in $\gL_2(f)$ with our estimate in Eq.~\ref{eq:importance-weight}, yielding a temporal difference (TD) InfoNCE estimator 
\begin{align}
    \mathcal{L}_{\text{TD InfoNCE}}(f) &\triangleq \mathbb{E}_{\substack{ (s, a) \sim p(s, a) \\ s_{t+}^{(2:N)} \sim p(s_{t+}) }} \left[ (1 - \gamma) \mathbb{E}_{s_{t+}^{(1)} \sim p(s' \mid s, a)} \left[ \log \frac{ e^{ f(s, a, s_{t+}^{(1)}) } }{\sum_{i = 1}^N e^{ f(s, a, s_{t+}^{(i)}) }} \right] \right. \nonumber \\
    & \hspace{3em} \left. + \gamma \mathbb{E}_{ \substack{s' \sim p(s' \mid s, a) \\ a' \sim \pi(a' \mid s') \\ s_{t+}^{(1)} \sim p(s_{t+}) }} \left[ \lfloor w(s', a', s_{t+}^{(1:N)}) \rfloor_{\text{sg}} \log \frac{ e^{ f(s, a, s_{t+}^{(1)}) } }{\sum_{i = 1}^N e^{ f(s, a, s_{t+}^{(i)}) }} \right] \right],
    \label{eq:td-infonce}
\end{align}
where $\lfloor \cdot \rfloor_{ \text{sg} }$ indicates the gradient of the importance weight should not affect the gradient of the entire objective. As shown in Fig.~\ref{fig:method}, we can interpret the first term as pulling together the representations of the current state-action pair $\phi(s, a)$ and the next state $\psi(s')$; the second term pulls the representations at the current step $\phi(s, a)$ similar to the (weighted) predictions from the future state $\psi(s_{t+})$. Importantly, the TD InfoNCE estimator is equivalent to the MC InfoNCE estimator for the optimal critic function: $\gL_{\text{TD InfoNCE}}(f^{\star}) = \gL_{\text{MC InfoNCE}}(f^{\star})$.

\paragraph{Convergence and connections.}
In Appendix~\ref{appendix:theoretical-analysis}, we prove that optimizing a variant of the TD InfoNCE objective is equivalent to perform one step policy evaluation with a new Bellman operator; thus, repeatedly optimizing this objective yields the correct discounted state occupancy measure. This analysis considers the tabular setting and assumes that the denominators of the softmax functions and $w$ in Eq.~\ref{eq:td-infonce} are computed using an exact expectation. We discuss the differences between TD InfoNCE and C-learning~\citep{eysenbach2020c} (a temporal difference estimator of the NCE objective) in Appendix~\ref{appendix:td-infonce-vs-c-learning}. Appendix~\ref{appendix:sr} discusses how TD InfoNCE corresponds to a nonparametric variant of the successor representation.

\subsection{Goal-conditioned Policy Learning}
\label{sec:alg}

The TD InfoNCE method provides a way for estimating the discounted state occupancy measure. This section shows how this estimator can be used to derive a new algorithm for goal-conditioned RL. This algorithm will alternate between \emph{(1)} estimating the occupancy measure using the TD InfoNCE objective and \emph{(2)} optimizing the policy to maximize the likelihood of the desired goal under the estimated occupancy measure. Pseudo-code is shown in Algorithm~\ref{alg:td-infonce}, and additional details are in Appendix~\ref{appendix:implementation}, and code is available online.%
\footnote{\url{https://github.com/chongyi-zheng/td_infonce}}

\begin{figure}[t]
\vspace{-3em}
\begin{algorithm}[H]
    \caption{Temporal Difference InfoNCE. We use $\CE$ to denote the cross entropy loss, taken across the rows of a matrix of logits and labels. We use $F$ as a matrix of logits, where $F[i, j] = \phi(s_t^{(i)}, a_t^{(i)}, g^{(i)})^{\top} \psi(s_{t+}^{(j)})$. See Appendix~\ref{appendix:implementation} for details.} 
    \label{alg:td-infonce}
    \begin{algorithmic}[1]
        \State{\textbf{Input}} contrastive representations $\phi_{\theta}$ and $\psi_{\theta}$, target representations $\phi_{\bar{\theta}}$ and $\psi_{\bar{\theta}}$, and goal-conditioned policy $\pi_{\omega}$.
        \For{each iteration}
            \State Sample $\{ (s_t^{(i)}, a_t^{(i)}, s_{t + 1}^{(i)}, g^{(i)}, s_{t+}^{(i)}) \}_{i = 1}^N \sim \text{replay buffer / dataset}, a^{(i)} \sim \pi(a \mid s_{t}^{(i)}, g^{(i)})$.
            \State Compute $F_{\text{next}}, F_{\text{future}}, F_{\text{goal}}$ using $\phi_{\theta}$ and $\psi_{\theta}$.
            \State Compute $\bar{F}_{w}$ using $\phi_{\bar{\theta}}$ and $\psi_{\bar{\theta}}$.
            \State $W \leftarrow N \cdot \texttt{stop\_grad} \left(\textsc{SoftMax}(\bar{F}_{w}) \right)$
            \State $\gL(\theta) \leftarrow (1 - \gamma) \CE(\text{logits} = F_{\text{next}}, \text{labels} = I_N) + \gamma \CE(\text{logits} = F_{\text{future}}, \text{labels} = W)$
            \State $\gL(\omega) \leftarrow \CE(\text{logits} = F_{\text{goal}}, \text{labels} = I_N)$
            \State Update $\theta, \omega$ by taking gradients of $\gL(\theta), \gL(\omega)$.
            \State Update $\bar{\theta}$ using an exponential moving average.
        \EndFor
        \State{\textbf{Return}} $\phi_{\theta}$, $\psi_{\theta}$, and $\pi_{\omega}$.
    \end{algorithmic}
\end{algorithm}
\vspace{-2em}
\end{figure}

While our TD InfoNCE loss in Sec.~\ref{subsec:td-infonce} estimates the discounted state occupancy measure for policy $\pi(a \mid s)$, we can extend it to the goal-conditioned setting by replacing $\pi(a \mid s)$ with $\pi(a \mid s, g)$ and $f(s, a, s_{t+})$ with $f(s, a, g, s_{t+})$, resulting in a goal-conditioned TD InfoNCE estimator. This goal-conditioned TD InfoNCE objective estimates the discounted state occupancy measure of~\emph{any} future state for a goal-conditioned policy commanding \emph{any} goal.
Recalling that the discounted state occupancy measure corresponds to the Q function~\citep{eysenbach2022contrastive}, the policy objective is to select actions that maximize the likelihood of the commanded goal:
\begin{align}
    &\mathbb{E}_{\substack{p_g(g),  p_0(s) \\ \pi(a_0 \mid s, g)}} \left[ \log p^{\pi}(s_{t+} = g \mid s, a, g) \right] = \mathbb{E}_{\substack{ g \sim p_g(g), s \sim p_0(s) \\ a_0 \sim \pi(a \mid s, g), s_{t+}^{(1:N)} \sim p(s_{t+}) }} \left[ \log \frac{e^{f^{\star}(s, a, g, s_{t+} = g )}}{ \sum_{i = 1}^N e^{ f^{\star}(s, a, g, s_{t+}^{(i)}) } } \right].
    \label{eq:actor-loss}
\end{align}
In practice, we optimize both the critic function and the policy for one gradient step iteratively, using our estimated $f$ in place of $f^{\star}$.

\section{Experiments}
\label{sec:experiments}

Our experiments start with comparing goal-conditioned TD InfoNCE to prior goal-conditioned RL approaches on both online and offline goal-conditioned RL (GCRL) benchmarks. We then analyze the properties of the critic function and the policy learned by this method. Visualizing the representations learned by TD InfoNCE reveals that linear interpolation corresponds to a form of planning. Appendix~\ref{appendix:td-infonce-vs-c-learning} ablates the difference between TD InfoNCE and a prior temporal difference method based on NCE. All experiments show means and standard deviations over five random seeds.

\subsection{Comparing to Prior Goal-conditioned RL methods}

We compare TD InfoNCE to four baselines on an online GCRL benchmark~\citep{plappert2018multi} containing four manipulation tasks for the Fetch robot. The observations and goals of those tasks can be either a state of the robot and objects or a $64 \times 64$ RGB image. We will evaluate using both versions. The first baseline, Quasimetric Reinforcement Learning (QRL)~\citep{pmlr-v202-wang23al}, is a state-of-the-art approach that uses quasimetric models to learn the optimal goal-conditioned value functions and the corresponding policies. The second baseline is contrastive RL~\citep{eysenbach2022contrastive}, which estimates the discounted state occupancy measure using $\gL_\text{MC InfoNCE}$ (Eq.~\ref{eq:mc-infonce}). Our third baseline is a variant of contrastive RL~\citep{eysenbach2022contrastive} using binary NCE loss. We call this method contrastive RL (NCE). The fourth baseline is the goal-conditioned behavioral cloning (GCBC)~\citep{ding2019goal, emmons2021rvs, ghosh2020learning, lynch2020learning, sun2019policy, srivastava2019training}. We also include a comparison with an off-the-shelf actor-critic algorithm augmented with hindsight relabeling~\citep{andrychowicz2017hindsight, levy2018learning, riedmiller2018learning, schaul2015universal} to learn a goal-conditioned policy (DDPG + HER).

\begin{figure}[t]
    \centering
    \vspace{-1.5em}
    \begin{subfigure}[c]{0.65\textwidth}
    \centering
        \includegraphics[width=\linewidth]{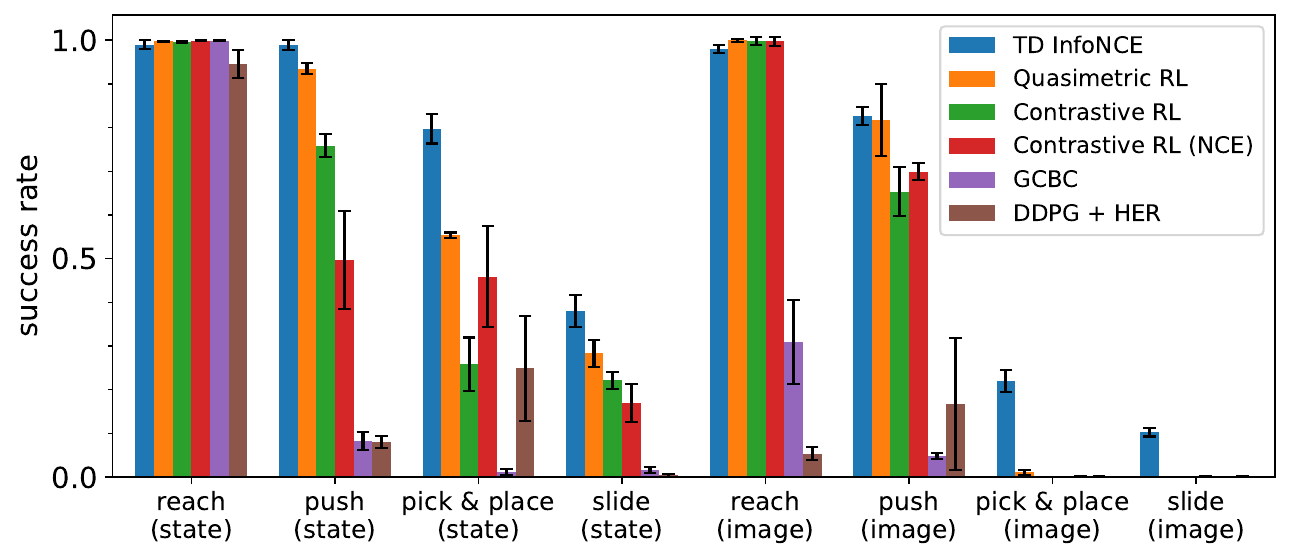}
        \caption{\footnotesize Fetch robotics benchmark from~\citep{plappert2018multi}}
        \label{fig:online-eval-bar}
    \end{subfigure}
    \hfill
    \begin{subfigure}[c]{0.34\textwidth}
        \centering
        \includegraphics[width=\linewidth]{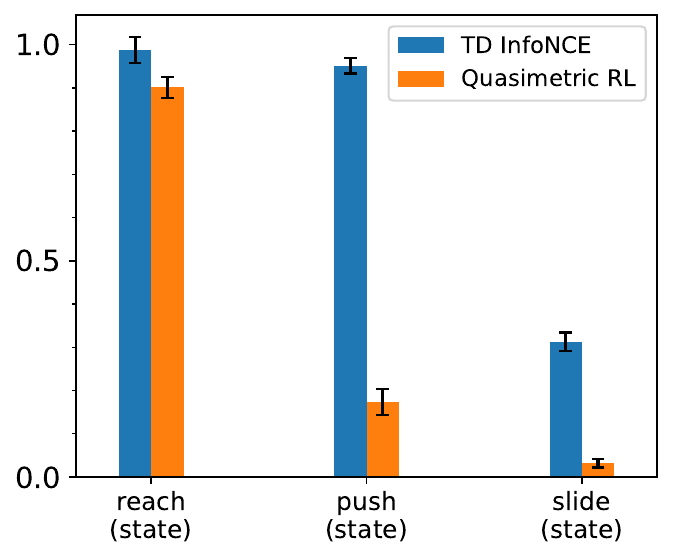}
        \caption{\footnotesize Stochastic tasks.}
        \label{fig:online-eval-stochastic-bar}
    \end{subfigure}
    \caption{\footnotesize \textbf{Evaluation on online GCRL benchmarks.} 
    \figleft \, TD InfoNCE performs similarly to or outperforms all baselines on both state-based and image-based tasks.
    \figright \, On stochastic versions of the state-based tasks, TD InfoNCE outperforms the strongest baseline (QRL).
    Appendix Fig.~\ref{fig:online-eval} shows the learning curves.
    }
    \vspace{-1em}
\end{figure}

We report results in Fig.~\ref{fig:online-eval-bar}, and defer the full learning curves to Appendix Fig.~\ref{fig:online-eval}. These results show that TD InfoNCE matches or outperforms other baselines on all tasks, both for state and image observations. On those more challenging tasks (\texttt{pick \& place (state / image)} and \texttt{slide (state / image)}), TD InfoNCE achieves a $2\times$ median improvement relative to the strongest baseline (Appendix Fig.~\ref{fig:online-eval}). On the most challenging tasks, image-based \texttt{pick \& place} and \texttt{slide}, TD InfoNCE is the only method achieving non-negligible success rates. For those tasks where the success rate fails to separate different methods significantly (e.g., \texttt{slide (state)} and \texttt{push (image)}), we include comparisons using minimum distances of the gripper or the object to the goal over an episode in Appendix Fig.~\ref{fig:online-eval-dist}. We speculate this observation is because TD InfoNCE estimates the discounted state occupancy measure more accurately, a hypothesis we will investigate in Sec.~\ref{subsec:critic-pred-acc}.

Among those baselines, QRL is the strongest one. Unlike TD InfoNCE, the derivation of QRL assumes the dynamics are deterministic.
This difference motivates us to study whether TD InfoNCE continues achieving high success rates in environments with stochastic noise. To study this, we compare TD InfoNCE to QRL on a variant of the Fetch benchmark where observations are corrupted with probability $0.1$. As shown in Fig.~\ref{fig:online-eval-stochastic-bar}, TD InfoNCE maintains high success rates while the performance of QRL decreases significantly, suggesting that TD InfoNCE can better cope with stochasticity in the environment.

\subsection{Evaluation on Offline Goal Reaching}

\begin{table}[t]
\caption{\footnotesize Evaluation on offline D4RL AntMaze benchmarks.}
\vspace{-1.5em}
\label{tab:offline-eval}
\begin{center}
\begin{small}
\setlength{\tabcolsep}{5pt}
\begin{tabular}{c|ccccccc}
\toprule
 & TD InfoNCE & QRL & Contrastive RL & GCBC & DT & IQL & TD3 + BC \\ 
\midrule
umaze-v2 & 84.9 $\pm$ 1.2 & $76.8 \pm 2.3$ & $79.8 \pm 1.6$ &  $65.4$ & $65.6$ & $\textbf{87.5}$ & $78.6$ \\
umaze-diverse-v2 & \textbf{91.7 $\pm$ 1.3} & $80.1 \pm 1.3$ & $77.6 \pm 2.8$ & $ 60.9$ & $51.2$ & $ 62.2$ & $71.4$ \\ 
medium-play-v2 & \textbf{86.8 $\pm$ 1.7} & $76.5 \pm 2.1$ & $72.6 \pm 2.9$ & $58.1$ & $1.0$ & $71.2$ & $10.6$ \\
medium-diverse-v2 & \textbf{82.0 $\pm$ 3.4} & $73.4 \pm 1.9$ & $71.5 \pm 1.3$ & $67.3$ & $0.6$ & $70.0$ & $3.0$ \\
large-play-v2 & $47.0 \pm 2.5$ & \textbf{52.9 $\pm$ 2.8} & $48.6 \pm 4.4$ & $32.4$ & $0.0$ & $39.6$ & $0.2$ \\
large-diverse-v2 & \textbf{55.6 $\pm$ 3.6} & $51.5 \pm 3.8$ & \textbf{54.1 $\pm$ 5.5} & $36.9$ & $0.2$ & $47.5$ & $0.0$ \\
\bottomrule
\end{tabular}
\end{small}
\end{center}
\end{table}

We next study whether the good performance of TD InfoNCE transfers to the setting without any interaction with the environment (i.e., offline RL). We evaluate on AntMaze tasks from the D4RL benchmark~\citep{fu2020d4rl}. The results in Table~\ref{tab:offline-eval} show that TD InfoNCE outperforms most baselines on most tasks. See Appendix~\ref{appendix:offline-details} for details.

\subsection{Accuracy of the estimated discounted state occupancy measure}
\label{subsec:critic-pred-acc}

\begin{figure*}[t]
    \centering
    \vspace{-0.5em}
    \includegraphics[width=\linewidth]{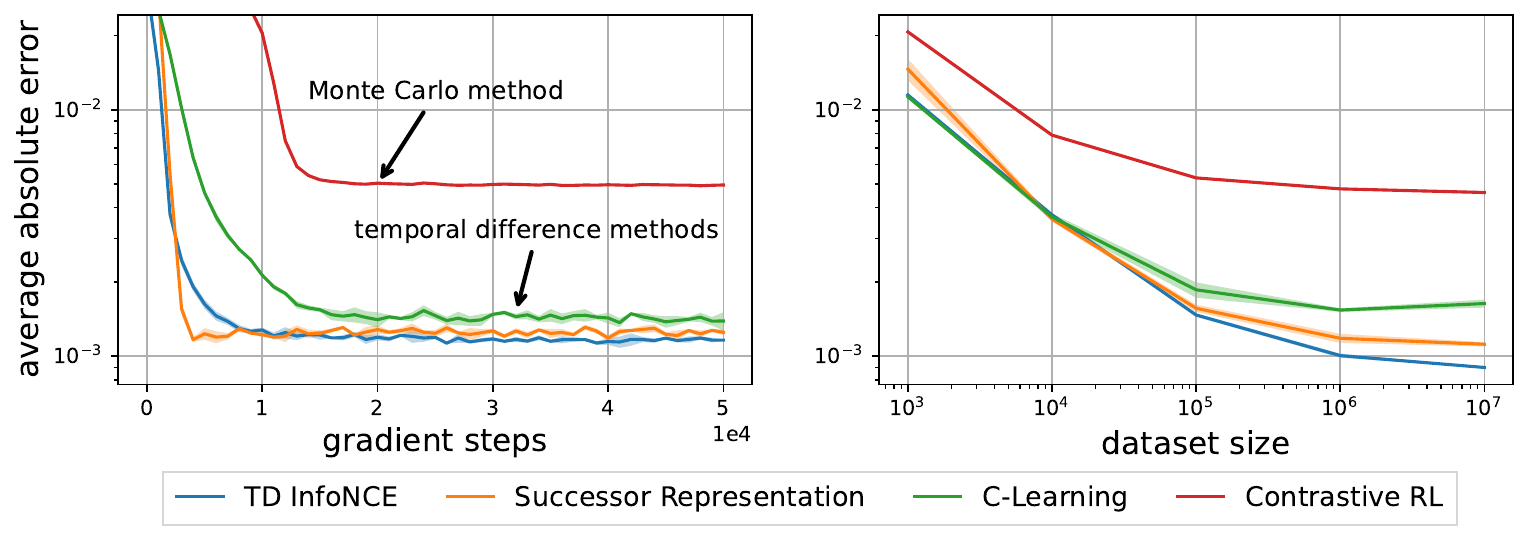}
    \caption{ \footnotesize \textbf{
    Estimating the discounted state occupancy measure in a tabular setting.} \figleft~Temporal difference methods have lower errors than the Monte Carlo method. Also note that our TD InfoNCE converges as fast as the best baseline (successor representation). \figright~TD InfoNCE is more data efficient than other methods. Using a dataset of size 10M, TD InfoNCE achieves an error rate $25\%$ lower than the best baseline; TD InfoNCE also matches the performance of C-learning with $130\times$ less data.
    }
    \label{fig:discounted-state-occupancy-measure-est-errs}
\end{figure*}

This section tests the hypothesis that our TD InfoNCE loss will be more accurate and sample efficient than alternative Monte Carlo methods (namely, contrastive RL~\citep{eysenbach2022contrastive}) in predicting the discounted state occupancy measure. We will use the tabular setting so that we can get a ground truth estimate.
We compare TD InfoNCE to three baselines.
Successor representations~\citep{dayan1993improving} can also be learned in a TD manner, though can be challenging to apply beyond tabular settings. C-learning is similar to TD InfoNCE in that it uses a temporal difference method to optimize a contrastive loss, but differs in using a binary cross entropy loss instead of a softmax cross entropy loss. Contrastive RL is the MC counterpart of TD InfoNCE.
We design a $5 \times 5$ gridworld with 125 states and 5 actions (up, down, left, right, and no-op) and collect 100K transitions using a uniform random policy, $\mu(a \mid s) = \textsc{Unif}(\gA)$. 
We evaluate each method by measuring the absolute error between the predicted probability $\hat{p}$ and the ground truth probability $p^{\mu}$, averaging over all pairs of $(s, a, s_{t+})$:
\begin{align*}
    \frac{1}{ \lvert \gS \rvert \lvert \gA \rvert \lvert \gS \rvert} \sum_{s, a, s_{t+}} \abs{ \hat{p}(s_{t+} \mid s, a) - p^{\mu}(s_{t+} \mid s, a)}.
\end{align*}
For the three TD methods, we compute the TD target in a SARSA manner~\citep{sutton2018reinforcement}. For those methods estimating a probability ratio, we convert the prediction to a probability by multiplying by the empirical state marginal. Results in Fig.~\ref{fig:discounted-state-occupancy-measure-est-errs} show that TD methods achieve lower errors than the Monte Carlo method, while TD InfoNCE converges faster than C-Learning. 
Appendix~\ref{appendix:critic-pred-acc-full} discusses why all methods plateau above zero.

Our next experiments studies sample efficiency. We hypothesize that the softmax in the TD InfoNCE loss may provide more learning signal than alternative methods, allowing it to achieve lower error on a fixed budget of data.
To test this hypothesis, we run experiments with dataset sizes from 1K to 10M on the same gridworld, comparing TD InfoNCE to the same set of baselines. We report results in Fig.~\ref{fig:discounted-state-occupancy-measure-est-errs} with errors showing one standard deviation after training for 50K gradient steps for each approach. These results suggest that methods based on temporal difference learning predict more accurately than Monte Carlo method when provided with the same amount of data.
Compared with its Monte Carlo counterpart, TD InfoNCE is $1500\times$ more sample efficient ($6.5 \times 10^3$ vs $10^7$ transitions).
Compared with the only other TD method applicable in continuous settings (C-learning), TD InfoNCE can achieve a comparable loss with $130\times$ less data ($7.7 \times 10^4$ vs $10^7$ transitions).
Even compared with the strongest baseline (successor representations), which makes assumptions (tabular MDPs) that our method avoids, TD InfoNCE can achieve a comparable error rate with almost $20 \times$ fewer samples ($5.2 \times 10^5$ vs $10^7$ transitions).

\subsection{Does TD InfoNCE enable off-policy reasoning?}
\label{subsec:off-policy-reasoning}

\begin{figure}[t]
    \centering
    \vspace{-1.5em}
    \begin{minipage}{0.48\textwidth}
        \begin{subfigure}[c]{0.32\linewidth}
        \centering
            \includegraphics[width=\linewidth]{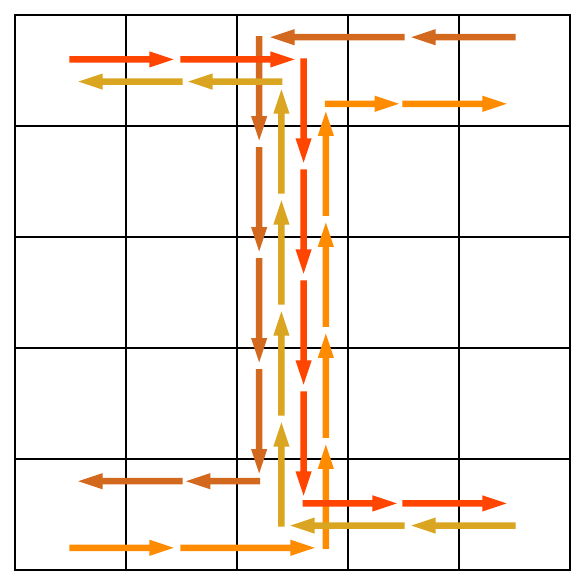}
            \caption*{Dataset}
        \end{subfigure}
        \begin{subfigure}[c]{0.32\linewidth}
            \includegraphics[width=\linewidth]{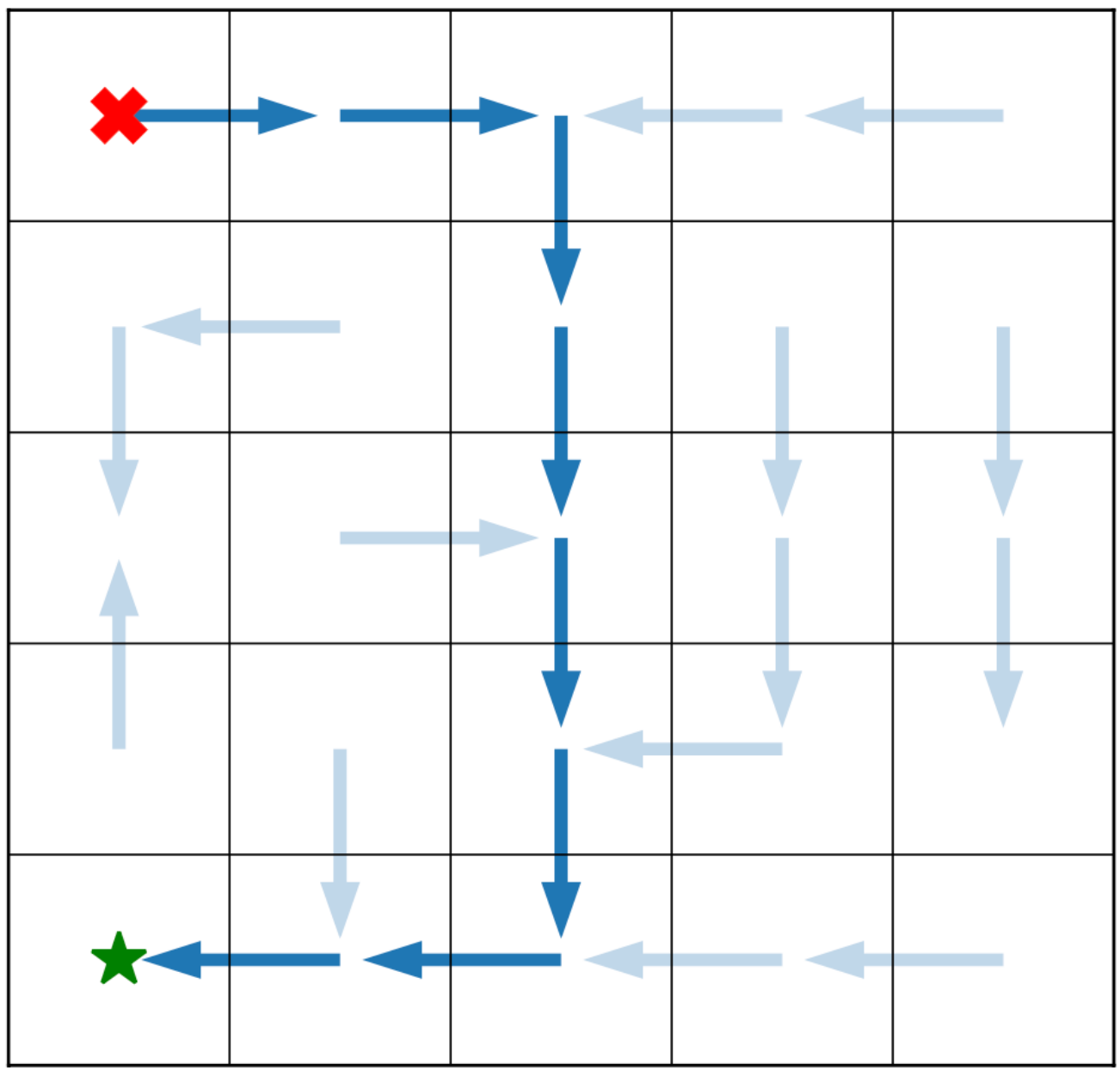}
            \caption*{TD InfoNCE}
        \end{subfigure}
        \begin{subfigure}[c]{0.32\linewidth}
            \includegraphics[width=\linewidth]{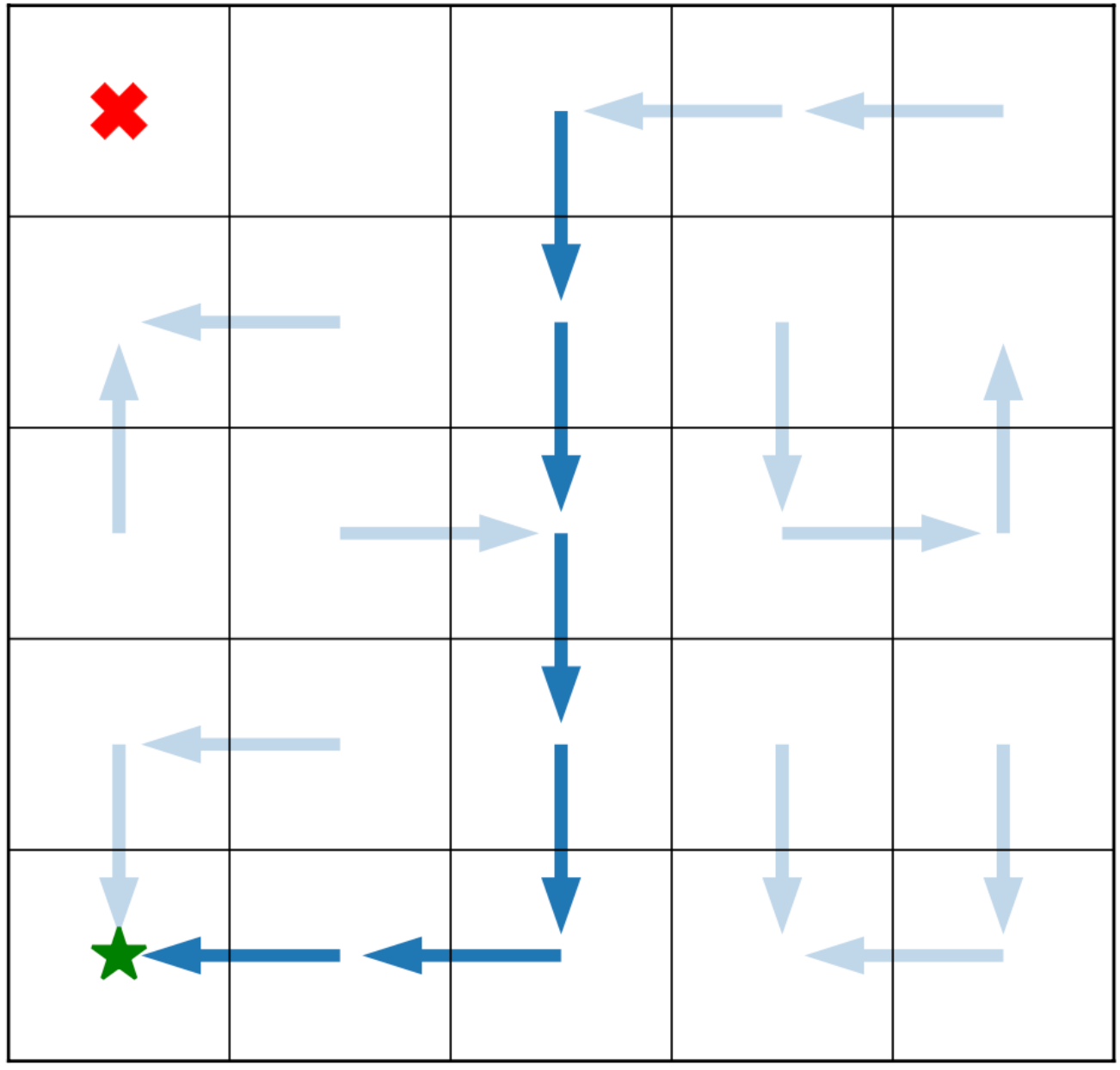}
            \caption*{Contrastive RL}
        \end{subfigure}
        \caption{\footnotesize \textbf{Stitching trajectories in a dataset.}  The behavioral policy collects ``Z" style trajectories. 
        Unlike the Monte Carlo method (contrastive RL) , our TD InfoNCE successfully ``stitches'' these trajectories together, navigating between pairs of (start \textcolor{Red}{\heavyxmark}, goal \textcolor{OliveGreen}{\starmark}) states unseen in the training trajectories. Appendix Fig.~\ref{fig:stitching-property-more} shows additional examples.
        }
        \label{fig:stitching-property}
    \end{minipage}
    \hfill
    \begin{minipage}{0.48\textwidth}
        \begin{subfigure}[c]{0.32\linewidth}
                \includegraphics[width=\linewidth]{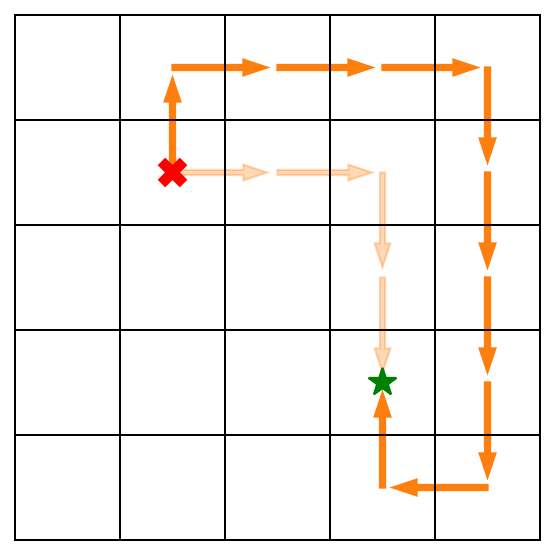}
                \caption*{Dataset}
            \end{subfigure}
            \hfill
            \begin{subfigure}[c]{0.32\linewidth}
                \includegraphics[width=\linewidth]{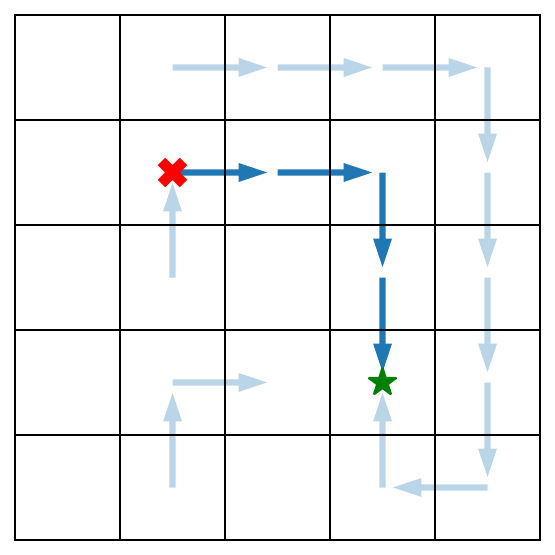}
                \caption*{TD InfoNCE}
            \end{subfigure}
            \hfill
            \begin{subfigure}[c]{0.32\linewidth}
                \includegraphics[width=\linewidth]{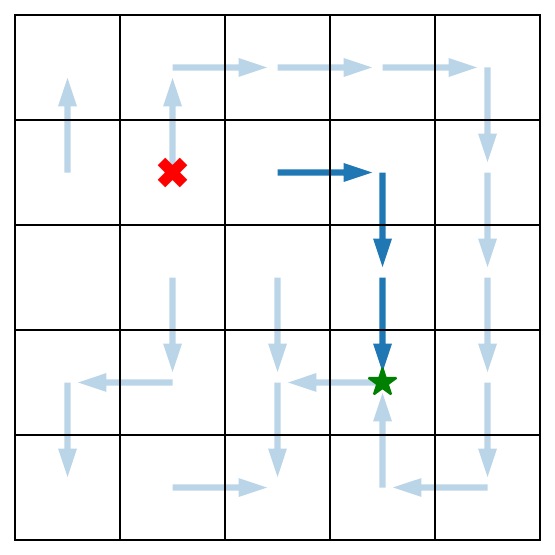}
                \caption*{Contrastive RL}
            \end{subfigure}
            \caption{\footnotesize \textbf{Searching for shortcuts in skewed datasets.}~\emph{(Left)} Conditioned on different initial states \textcolor{Red}{\heavyxmark} and goals \textcolor{OliveGreen}{\starmark}, we collect datasets with $95\%$ long paths (dark) and $5\%$ short paths (light).~\emph{(Center)} TD InfoNCE infers the shortest path, \emph{(Right)} while contrastive RL fails to find this path. Appendix Fig.~\ref{fig:searching-shortcut-more} shows additional examples.}
            \label{fig:searching-shotcut}
    \end{minipage}
\end{figure}

The explicit temporal difference update (Eq.~\ref{eq:td-infonce}) in TD InfoNCE is similar to the standard Bellman backup, motivating us to study whether the resulting goal-conditioned policy is capable of performing dynamic programming with offline data. To answer these questions, we conduct two experiments on the same gridworld environment as in Sec.~\ref{subsec:critic-pred-acc}, comparing TD InfoNCE to contrastive RL (i.e., Monte Carlo InfoNCE).
Fig.~\ref{fig:stitching-property} shows that TD InfoNCE successfully stitches together pieces of different trajectories to find a route between unseen (state, goal) pairs. Fig.~\ref{fig:searching-shotcut} shows that TD InfoNCE can perform off-policy reasoning, finding a path that is shorter than the average path demonstrated in the dataset. See Appendix~\ref{appendix:off-policy} for details.

\section{Conclusion}
\label{sec:conclusion}

This paper introduced a temporal difference estimator for the InfoNCE loss. Our goal-conditioned RL algorithm based on this estimator outperforms prior methods in both online and offline settings, and is capable of handling stochasticity in the environment dynamics.
While we focused on a specific type of RL problem (goal-conditioned RL), in principle the TD InfoNCE estimator can be used to drive policy evaluation for arbitrary reward functions. One area for future work is to determine how it compares to prior off-policy evaluation techniques.

While we focused on evaluating the TD InfoNCE estimator on control tasks, it is worth noting that the MC InfoNCE objective has been previously applied to NLP, audio, video settings; one intriguing and important question is whether the benefits of TD learning seen on these control tasks translate into better representations in these other domains.

\paragraph{Limitations.} One limitation of TD InfoNCE is complexity: compared with its Monte Carlo counterpart, ours is more complex and requires more hyperparameters. It is worth noting that even TD InfoNCE struggles to solve the most challenging control tasks with image observations. On the theoretical front, our convergence proof uses a slightly modified version of our loss (replacing a sum with an expectation), which would be good to resolve in future work.

\section*{Acknowledgements}

We thank Ravi Tej and Wenzhe Li for discussions about this work, and anonymous reviewers for providing feedback on early versions of this work. We thank Tongzhou Wang for providing performance of baselines in online GCRL experiments and thank Raj Ghugare for sharing code of environment implementation. We thank Vivek Myers for finding issues in the code. BE is supported by the Fannie and John Hertz Foundation.

\clearpage
\appendix

\section{Theoretical Analysis}
\label{appendix:theoretical-analysis}

Our convergence proof will focus on the tabular setting with known $p(s' \mid s, a)$ and $p(s_{t+})$, and follows the fitted Q-iteration strategy~\citep{fu2019diagnosing, ernst2005tree, bertsekas1995neuro}: at each iteration, an optimization problem will be solved exactly to yield the next estimate of the discounted state occupancy measure. One key step in the proof is to employ a preserved invariant; we will define the classifier derived from the TD InfoNCE objective (Eq.~\ref{eq:td-infonce}) and show that this classifier always represents a valid probability distribution (over future states).
We then construct a variant of the TD InfoNCE objective using this classifier and prove that optimizing this objective is exactly equivalent to perform policy evaluation, resulting in the convergence to the discounted state occupancy measure.

\paragraph{Definition of the classifier.} We start by defining the classifier derived from the TD InfoNCE as
\begin{align}
    C(s, a, s_{t+}) \triangleq \frac{p(s_{t+}) e^{f(s, a, s_{t+})}}{ \E_{p(s_{t+}')}\left[ e^{f(s, a, s_{t+}')} \right]} = \frac{p(s_{t+}) e^{f(s, a, s_{t+})}}{ \sum_{s_{t+}' \in \gS} p(s_{t+}) e^{f(s, a, s_{t+}')}},
    \label{eq:classifier}
\end{align}
suggesting that $C(s, a, \cdot)$ is a valid distribution over future states: $C(s, a, \cdot) \in \Delta(\gS)$. 

\paragraph{A variant of TD InfoNCE.} Our definition of the classifier (Eq.~\ref{eq:classifier}) allows us to rewrite the importance weight $w(s, a, s_{t+})$ and softmax functions in $\gL_{\text{TD InfoNCE}}$ (Eq.~\ref{eq:td-infonce}) as Monte Carlo estimates of the classifier using samples of $s_{t+}^{(1:N)}$,
\begin{align*}
    w(s, a, s_{t+}^{(1:N)}) &= \frac{e^{f(s, a, s_{t+}^{(1)})}}{ \frac{1}{N} \sum_{i = 1}^N e^{f(s, a, s_{t+}^{(i)})}} \approx \frac{C(s, a, s_{t+})}{p(s_{t+})}.
\end{align*}
Thus, we construct a variant of the TD InfoNCE objective using $C$:
\begin{align*}
    \bar{\gL}_{\text{TD InfoNCE}}(C) &\triangleq \E_{p(s, a)} \left[ (1 - \gamma) \E_{p(s' = s_{t+} \mid s, a)} \left[\log C(s, a, s_{t+}) \right] \right. \nonumber \\ 
    &\hspace{4.5em} \left. + \gamma \E_{\substack{p(s' \mid s, a), \pi(a' \mid s') \\ p(s_{t+}) } } \left[ \frac{\lfloor C(s', a', s_{t+}) \rfloor_{\text{sg}}}{p(s_{t+})} \log C(s, a, s_{t+}) \right] \right].
\end{align*}
This objective is similar to $\gL_{\text{TD InfoNCE}}$, but differs in that \emph{(a)} softmax functions are replaced by $C(s, a, s_{t+})$ up to constant $\frac{1}{N \cdot p(s_{t+})}$ and \emph{(b)} $w(s', a', s_{t+}^{(1:N)})$ is replaced by $\frac{C(s', a', s_{t+})}{p(s_{t+})}$. Formally, $\gL_{\text{TD InfoNCE}}(C)$ is a nested Monte Carlo estimator of $\bar{\gL}_{\text{TD InfoNCE}}$~\citep{rainforth2018nesting,giles2015multilevel} and we leave the analysis of the gap between them as future works. We now find the solution of $\bar{\gL}_{\text{TD InfoNCE}}(C)$ analytically by rewriting it using the cross entropy and ignore the stop gradient operator to reduce clutter: $\bar{\gL}_{\text{TD InfoNCE}}(C) =$
\begin{align}
    &\E_{p(s, a)} \left[ (1 - \gamma) \E_{p(s' = s_{t+} \mid s, a)} \left[\log C(s, a, s_{t+}) \right] + \gamma \E_{\substack{p(s' \mid s, a), \pi(a' \mid s', g) \\ C(s', a', s_{t+}) } } \left[ \log C(s, a, s_{t+}) \right] \right] \nonumber \\
    &= -\E_{p(s, a)} \left[ (1 - \gamma) \CE \left(p(s' = \cdot \mid s, a), C(s, a, \cdot) \right) \right. \nonumber \\
    & \hspace{5em} \left. + \gamma \CE \left(\E_{p(s' \mid s, a), \pi(a' \mid s')} \left[ C(s', a', \cdot) \right], C(s, a, \cdot) \right) \right] \nonumber \\
    &= -\E_{p(s, a)} \left[ \CE \left((1 - \gamma) p(s' = \cdot \mid s, a) + \gamma \E_{p(s' \mid s, a), \pi(a' \mid s')} \left[ C(s', a', \cdot) \right], C(s, a, \cdot) \right) \right],
    \label{eq:td-infonce-expectation-ce}
\end{align}
where the cross entropy for $p, q \in \Delta(\gX)$ is defined as 
\begin{align*}
    \CE(p(\cdot), q(\cdot)) = - \E_{p(x)}[\log q(x)] = - \sum_{x \in \gX} p(x) \log q(x),
\end{align*}
with the minimizer $q^{\star} = \argmin_{q \in \Delta(\gX)} \CE(p(\cdot), q(\cdot)) = p$. Note that $p(s' = \cdot \mid s, a) \in \Delta(\gS)$ and $\E_{p(s' \mid s, a)\pi(a' \mid s')}[C(s', a', \cdot)] \in \Delta(\gS)$ in Eq.~\ref{eq:td-infonce-expectation-ce} indicate that their convex combination is also a distribution in $\Delta(\gS)$. This objective suggests a update for the classifier given any $(s, a, s_{t+})$:
\begin{align}
    C(s, a, s_{t+}) \gets (1 - \gamma) p(s' = s_{t+} \mid s, a) + \gamma \E_{p(s' \mid s, a) \pi(a' \mid s')}[C(s', a', s_{t+})],
    \label{eq:td-infonce-update}
\end{align}
which bears a resemblance to the standard Bellman equation. 

\paragraph{InfoNCE Bellman operator.} We define the InfoNCE Bellman operator for any function $Q(s, a, s_{t+}): \gS \times \gA \times \gS \mapsto \R$ with policy $\pi(a \mid s)$ as 
\begin{align}
    \gT_{\text{InfoNCE}} Q(s, a, s_{t+}) \triangleq (1 - \gamma) p(s' = s_{t+} \mid s, a) + \gamma \E_{p(s' \mid s, a) \pi(a' \mid s')}[Q(s', a', s_{t+})],
\end{align}
and write the update of the classifier as $C(s, a, s_{t+}) \gets \gT_{\text{InfoNCE}} C(s, a, s_{t+})$. Like the standard Bellman operator, this InfoNCE Bellman operator is a $\gamma$-contraction. Unlike the standard Bellman operator, $\gT_{\text{InfoNCE}}$ replaces the reward function with the discounted probability of the future state being the next state $(1 - \gamma) p(s' = s_{t+} \mid s, a)$ and applies to a function depending on a state-action pair and a future state $(s, a, s_{t+})$.

\paragraph{Proof of convergence.} Using the same proof of convergence for policy evaluation with the standard Bellman equation~\citep{sutton2018reinforcement, agarwal2019reinforcement}, we conclude that repeatedly applying $\gT_{\text{InfoNCE}}$ to $C$ results in convergence to a unique $C^{\star}$ regardless of initialization,
\begin{align*}
    C^{\star}(s, a, s_{t+}) = (1 - \gamma) p(s' = s_{t+} \mid s, a) + \gamma \E_{p(s' \mid s, a) \pi(a' \mid s')}[C^{\star}(s', a', s_{t+})].
\end{align*}
Since $C^{\star}(s, a, s_{t+})$ and $p^{\pi}(s_{t+} \mid s, a)$ satisfy the same identity (Eq.~\ref{eq:discounted-state-occupancy-measure-recurrence}), we have $C^{\star}(s, a, s_{t+}) = p^{\pi}(s_{t+} \mid s, a)$, i.e., the classifier of the TD InfoNCE estimator converges to the discounted state occupancy measure. To recover $f^{\star}$ from $C^{\star}$, we note that $f^{\star}$ satisfies
\begin{align*}
    f^{\star}(s, a, s_{t+}) &= \log C^{\star}(s, a, s_{t+}) - \log p(s_{t+}) + \log \E_{p(s_{t+}')}[\exp(f^{\star}(s, a, s_{t+}'))] \\
    &= \log p^{\pi}(s_{t+} \mid s, a) - \log p(s_{t+}) + \log \E_{p(s_{t+}')}[\exp(f^{\star}(s, a, s_{t+}'))]
\end{align*}
by definition.
Since the (expected) softmax function is invariant to translation, we can write $f^{\star}(s, a, s_{t+}) = \log p^{\pi}(s_{t+} \mid s, a) - \log p(s_{t+}) - \log c(s, a)$, where $c(s, a)$ is an arbitrary function that does not depend on $s_{t+}$~\footnote{Technically, $f^{\star}$ should be a set of functions satisfying $\left\{ f: \frac{e^{f(s, a, s_{t+})}}{\E_{p(s_{t+}')}\left[e^{f(s, a, s_{t+}')} \right]} = \frac{C^{\star}(s, a, s_{t+})}{p(s_{t+})} \right\}.$}. Thus, we conclude that TD InfoNCE objective converges to the same solution as that of MC InfoNCE (Eq.~\ref{eq:opt-critic}), i.e. $\bar{\gL}_{\text{TD InfoNCE}}(f^{\star}) = \gL_{\text{MC InfoNCE}}(f^{\star})$.

It is worth noting that the same proof applies to the goal-conditioned TD InfoNCE objective. After finding an exact estimate of the discounted state occupancy measure of a goal-conditioned policy $\pi(a \mid s, g)$, maximizing the policy objective (Eq.~\ref{eq:actor-loss}) is equivalent to doing policy improvement. We can apply the same proof as in the Lemma 5 of~\citep{eysenbach2020c} to conclude that $\pi(a \mid s, g)$ converges to the optimal goal-conditioned policy $\pi^{\star}(a \mid s, g)$.

\section{Connection with mutual information and skill learning.}
\label{appendix:mi}

The theoretical analysis in Appendix~\ref{appendix:theoretical-analysis} has shown that the TD InfoNCE estimator has the same effect as the MC InfoNCE estimator. As the (MC) InfoNCE objective corresponds to a lower bound on mutual information~\citep{poole2019variational}, we can interpret our goal-conditioned RL method as having both the actor and the critic jointly optimize a lower bound on mutual information. This perspective highlights the close connection between unsupervised skill learning algorithms~\citep{eysenbach2018diversity, campos2020explore, warde2018unsupervised, gregor2016variational}, and goal-conditioned RL, a connection previously noted in~\citet{choi2021variational}. Seen as an unsupervised skill learning algorithm, goal-conditioned RL lifts one of the primary limitations of prior methods: it can be unclear which skill will produce which behavior. In contrast, goal-conditioned RL methods learn skills that are defined as optimizing the likelihood of reaching particular goal states.

\section{Connection with Successor Representations}
\label{appendix:sr}

In settings with tabular states, the successor representation~\citep{dayan1993improving} is a canonical method for estimating the discounted state occupancy measure (Eq.~\ref{eq:discounted-state-occupancy-measure}). The successor representation has strong ties to cognitive science~\citep{gershman2018successor} and has been used to accelerate modern RL methods~\citep{barreto2017successor, touati2021learning}.

Successor representation $M^{\pi}: \mathcal{S} \times \mathcal{A} \mapsto \Delta(\mathcal{S})$ is a long-horizon, policy dependent model that estimates the discounted state occupancy measure for every $s \in \mathcal{S}$ via the recursive relationship (Eq.~\ref{eq:discounted-state-occupancy-measure-recurrence}). Given a policy $\pi(a \mid s)$, the successor representation satisfies
\begin{align}
    M^{\pi}(s, a) \gets (1 - \gamma) \textsc{OneHot}_{|\mathcal{S}|}(s') + \gamma M^{\pi}(s', a'),
    \label{eq:sr}
\end{align}
where $s' \sim p(s' \mid s, a)$ and $a' \sim \pi(a' \mid s')$.
Comparing this update to the TD InfoNCE update shown in Fig.~\ref{fig:method} and Eq.~\ref{eq:td-infonce-update}, we see that this successor representation update is a special case of TD InfoNCE where \emph{(a)} every state is used instead of randomly-sampling the states, and \emph{(b)} the probabilities are encoded directed in a matrix $M$, rather than encoding the probabilities as the inner product between two learned vectors.

This connection is useful because it highlights how and why the learned representations can be used to solve fully-general reinforcement learning tasks. In the same way that the successor representation can be used to express the value function of a reward ($M^\pi(s, a)^\top r(\cdot)$), the representations learned by TD InfoNCE can be used to recover value functions:
\begin{align*}
    \hat{Q}^\pi(s, a) &= r(s, a) + \frac{\gamma}{1 - \gamma} \E_{s_{t+}^{(1:N)} \sim p(s_{t+}), a_{t+} \sim \pi(a \mid s_{t+}^{(1)})} \left[ \frac{e^{f(s, a, s_{t+}^{(1)})}}{ \frac{1}{N} \sum_{i=1}^N e^{f(s, a, s_{t+}^{(i)})}} r(s_{t+}^{(1)}, a_{t+}) \right]
\end{align*}
See~\citet{mazoure2022contrastive} for details on this construction.

\section{Experimental Details}

\subsection{The Complete Algorithm for Goal-Conditioned RL}
\label{appendix:implementation}

The complete algorithm of TD InfoNCE alters between estimating the discounted state occupancy measure of the current goal-conditioned policy via contrastive learning (Eq.~\ref{eq:td-infonce}) and updating the policy using the actor loss (Eq.~\ref{eq:actor-loss}), while collecting more data. Given a batch of $N$ transitions of $\{ (s_t^{(i)}, a_t^{(i)}, s_{t + 1}^{(i)}, g^{(i)}, s_{t+}^{(i)}) \}_{i = 1}^N$ sampled from $p(s_t, a_t, g)$, $p(s_{t + 1} \mid s_t, a_t)$, and $p(s_{t+})$, we can first compute the critic function for different combinations of goal-conditioned state-action pairs and future states by computing their contrastive representations $\phi(s_t, a_t, g)$, $\psi(s_{t+})$, and $\psi(s_{t+})$, and then construct two critic matrices $F_{\text{next}}, F_{\text{future}} \in \mathbb{R}^{N \times N}$ using those representations:
\begin{align*}
    F_{\text{next}}[i, j] = \phi(s_t^{(i)}, a_t^{(i)}, g^{(i)})^{\top} \psi(s_{t + 1}^{(j)}), F_{\text{future}}[i, j] = \phi(s_t^{(i)}, a_t^{(i)}, g^{(i)})^{\top} \psi(s_{t+}^{(j)})
\end{align*}
Note that the inner product parameterization of the critic function $f(s_t, a_t, g, s_{t+}) = \phi(s_t, a_t, g)^{\top} \psi(s_{t+})$ helps compute these matrices efficiently. Using these critic matrices, we rewrite the TD InfoNCE estimate as a sum of two cross entropy losses. The first cross entropy loss involves predicting which of the $N$ next states $s_{t + 1}^{(1:N)}$ is the correct next state for the corresponding goal-conditioned state and action pair:
\begin{align*}
    (1 - \gamma) \CE(\text{logits} = F_{\text{next}}, \text{labels} = I_N),
\end{align*}
where $\CE(\text{logits} = F_{\text{next}}, \text{labels} = I_N) = -\sum_{i = 1}^N \sum_{j = 1}^N I_N[i, j] \cdot \log \textsc{SoftMax}(F_{\text{next}})[i, j]$, $\textsc{SoftMax}(\cdot)$ denotes row-wise softmax normalization, and $I_N$ is a $N$ dimensional identity matrix. For the second cross entropy term, we first sample a batch of $N$ actions from the target policy at the~\emph{next} time step, $a_{t + 1}^{(1:N)} \sim \pi(a_{t + 1} \mid s_{t + 1}, g)$, and then estimate the importance weight matrix $W \in \mathbb{R}^{N \times N}$ that serves as labels as
\begin{align*}
    F_w[i, j] = \phi(s_{t + 1}^{(i)}, a_{t + 1}^{(i)}, g^{(i)})^{\top} \psi(s_{t+}^{(j)}), W = N \cdot \textsc{SoftMax}(F_{w}).
\end{align*}
Thus, the second cross entropy loss takes as inputs the critic $F_{\text{future}}$ and the importance weight $W$:
\begin{align}
    \gamma \CE(\text{logits} = F_{\text{future}}, \text{labels} = W).
    \label{eq:ce-negative}
\end{align}
Regarding the policy objective (Eq.~\ref{eq:actor-loss}), it can also be rewritten as the cross entropy between a critic matrix $F_{\text{goal}}$ with $F_{\text{goal}}[i, j] = \phi(s_t^{(i)}, {a}^{(i)}, g^{(i)})^{\top} \psi(g^{(j)})$, where $a^{(i)} \sim \pi(a \mid s_t^{(i)}, g^{(i)})$, and the identity matrix $I_N$:
\begin{align*}
    \CE(\text{logits} = F_{\text{goal}}, \text{labels} = I_N)
\end{align*}
In practice, we use neural networks with parameters $\theta = \{ \theta_{\phi}, \theta_{\psi} \}$ to parameterize (normalized) contrastive representations $\phi$ and $\psi$ and use a neural network with parameters $\omega$ to parameterize the goal-conditioned policy $\pi$ and optimize them using gradient descent.

\subsection{Online Goal-conditioned RL Experiments}

\begin{figure*}[t]
    \centering
    \begin{subfigure}[c]{\textwidth}
    \centering
        \includegraphics[width=\linewidth]{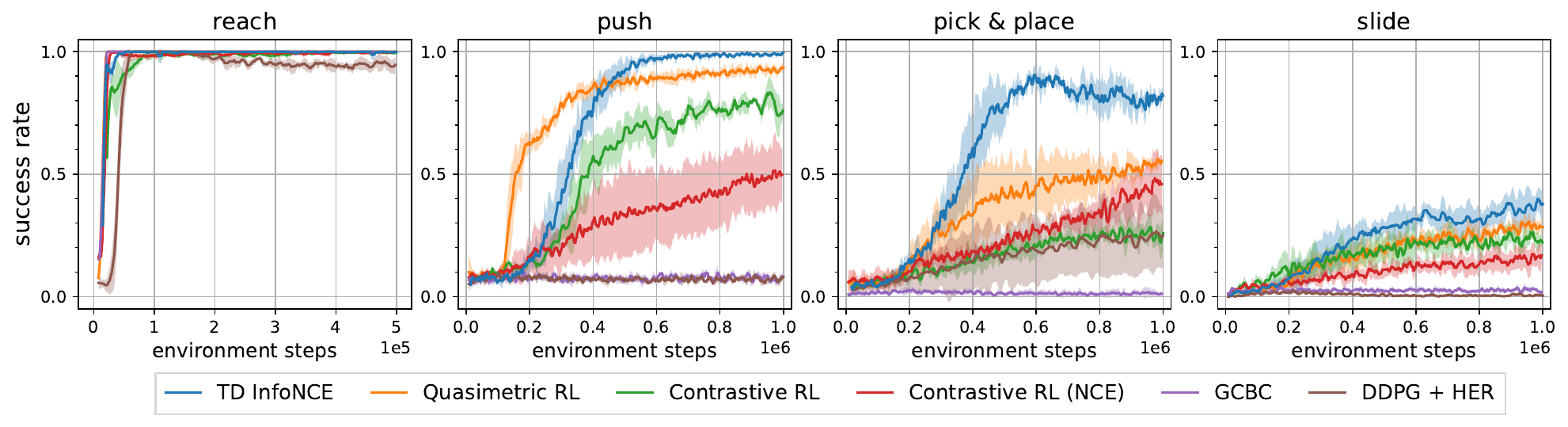}
        \caption{State-based tasks}
    \end{subfigure}
    \vfill
    \begin{subfigure}[c]{\textwidth}
        \centering
        \includegraphics[width=\linewidth]{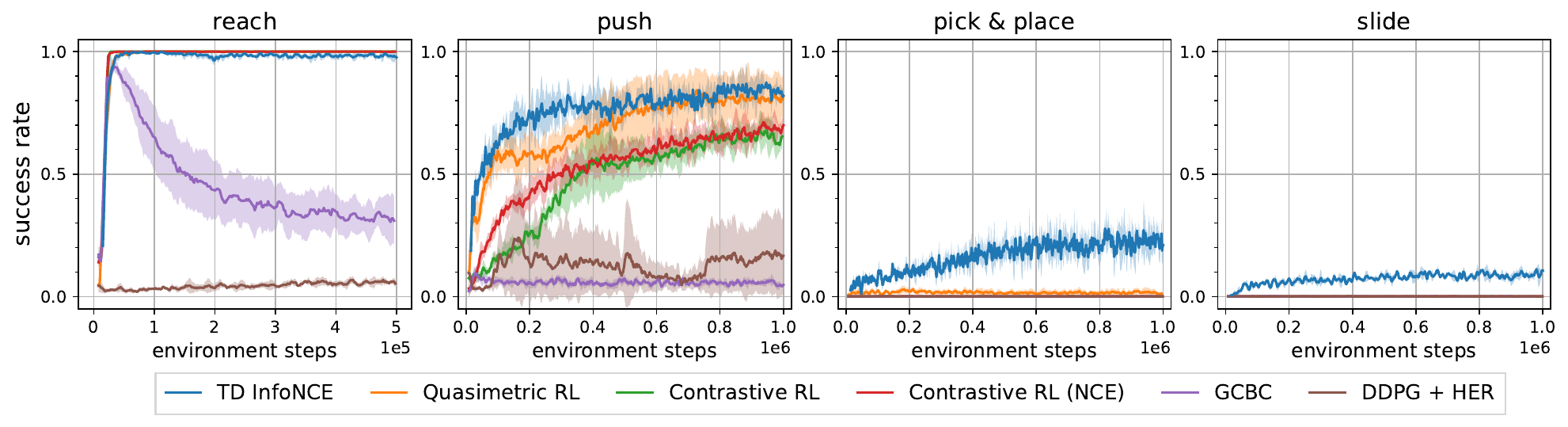}
        \caption{Image-based tasks}
    \end{subfigure}
    \caption{ \textbf{Evaluation on online GCRL benchmarks.} TD InfoNCE matches or outperforms all baselines on both state-based and image-based tasks.}
    \label{fig:online-eval}
\end{figure*}

\begin{figure*}[t]
    \centering
    \includegraphics[width=\linewidth]{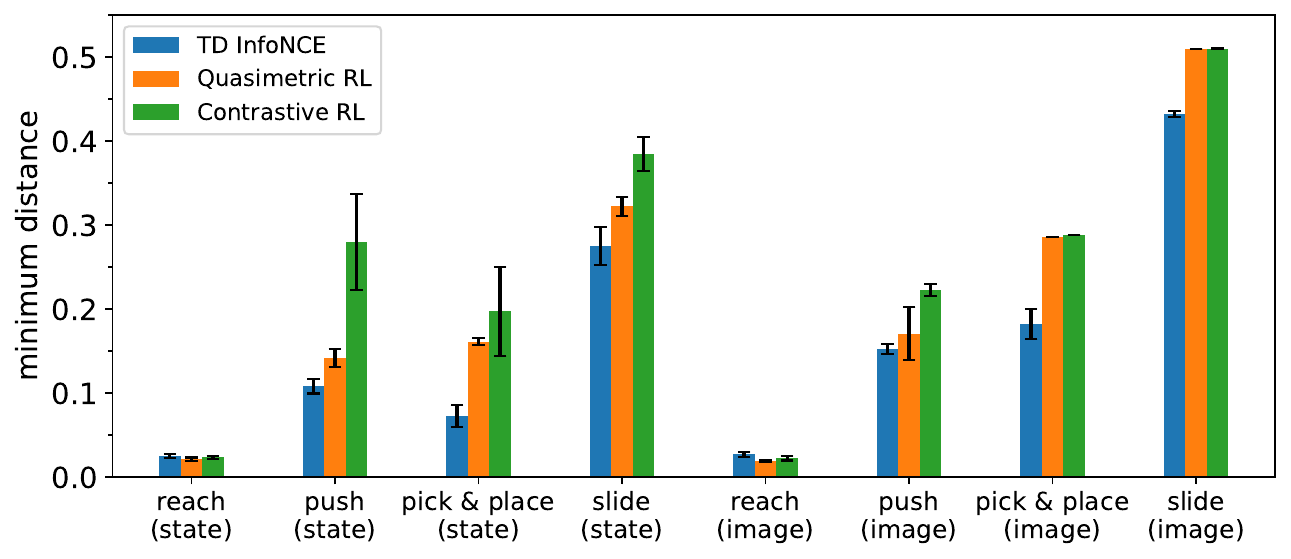}
    \caption{We also compare different methods using the minimum distance of the gripper or the object to the goal over an episode. Note that a lower minimum distance indicates a better performance. TD InfoNCE achieves competitive minimum distances on online GCRL benchmarks.}
    \label{fig:online-eval-dist}
\end{figure*}

We report complete success rates for online GCRL experiments in Fig.~\ref{fig:online-eval}, showing the mean success rate and standard deviation (shaded regions) across five random seeds. TD InfoNCE outperforms or achieves similar performance on all tasks, compared with other baselines. For those tasks where the success rate fails to separate different methods significantly (e.g., \texttt{slide (state)} and \texttt{push (image)}), we include comparisons using minimum distances of the gripper or the object to the goal over an episode in Fig.~\ref{fig:online-eval-dist}, selecting the strongest baselines QRL and contrastive RL. Note that a lower minimum distance indicates a better performance. These results suggest that TD InfoNCE is able to emerge a goal-conditioned policy by estimating the discounted state occupancy measure, serving as a competitive goal-conditioned RL algorithm.

\subsection{Offline Goal-conditioned RL Experiments}
\label{appendix:offline-details}

Similar to prior works~\citep{eysenbach2022contrastive, pmlr-v202-wang23al}, we adopt an additional goal-conditioned behavioral cloning regularization to prevent the policy from sampling out-of-distribution actions~\citep{fujimoto2021minimalist, kumar2020conservative, kumar2019stabilizing} during policy optimization (Eq.\ref{eq:policy-obj}):
\begin{align*}
    \argmax_{\pi(\cdot \mid \cdot, \cdot)} \mathbb{E}_{\substack{ (s, a_{\text{orig}}, g) \sim p(s, a_{\text{orig}}, g) \\ a \sim \pi(a \mid s, g), s_{t+}^{(1:N)} \sim p(s_{t+}) }} \left[ (1 - \lambda) \cdot \log \frac{e^{f(s, a, g, s_{t+} = g )}}{ \sum_{i = 1}^N e^{ f(s, a, g, s_{t+}^{(i)}) } } + \lambda \cdot \|a - a_{\text{orig}}\|_2^2 \right],
\end{align*}
where $\lambda$ is the coefficient for regularization. Note that we use a supervised loss based on the mean squared error instead of the maximum likelihood estimate of $a_{\text{orig}}$ under policy $\pi$ used in prior works. We compare TD InfoNCE to the state-of-the-art QRL~\citep{pmlr-v202-wang23al} and its Monte Carlo counterpart (contrastive RL~\citep{eysenbach2022contrastive}). We also compare to the pure goal-conditioned behavioral cloning implemented in~\citep{emmons2021rvs} as well as a recent baseline that predicts optimal actions via sequence modeling using a transformer (DT~\citep{chen2021decision}). Our last two baselines are offline actor-critic methods trained via TD learning: TD3 + BC~\citep{fujimoto2021minimalist} and IQL~\citep{kostrikov2021offline}, not involving goal-conditioned relabeling. We use the result for baselines except QRL from~\citep{eysenbach2022contrastive}.

As shown in Table~\ref{tab:offline-eval}, TD InfoNCE matches or outperforms all baselines on 5 / 6 tasks. On tasks (\texttt{medium-play-v2} and \texttt{medium-diverse-v2}), TD InfoNCE achieves a $+13\%$ improvement over contrastive RL, showing the advantage of temporal difference learning over the Monte Carlo approach with a fixed dataset. We conjecture that this benefit comes from the dynamic programming property of the TD method and will investigate this property further in later experiments (Sec.~\ref{subsec:off-policy-reasoning}). Additionally, TD InfoNCE performs $1.4\times$ better than GCBC and retains a $3.8\times$ higher scores than DT on average, where these baselines use (autoregressive) supervised losses instead of TD learning. These results suggest that TD InfoNCE is also a competitive goal-conditioned RL algorithm in the offline setting.

\subsection{Off-Policy Reasoning Experiments}
\label{appendix:off-policy}

\begin{figure*}
    \centering
    \begin{subfigure}[c]{0.24\textwidth}
    \centering
        \includegraphics[width=\linewidth]{figures/policy_analysis/stitching_dataset.pdf}
        Dataset
    \end{subfigure}
    \hfill
    \begin{subfigure}[c]{0.75\textwidth}
        \includegraphics[width=\linewidth]{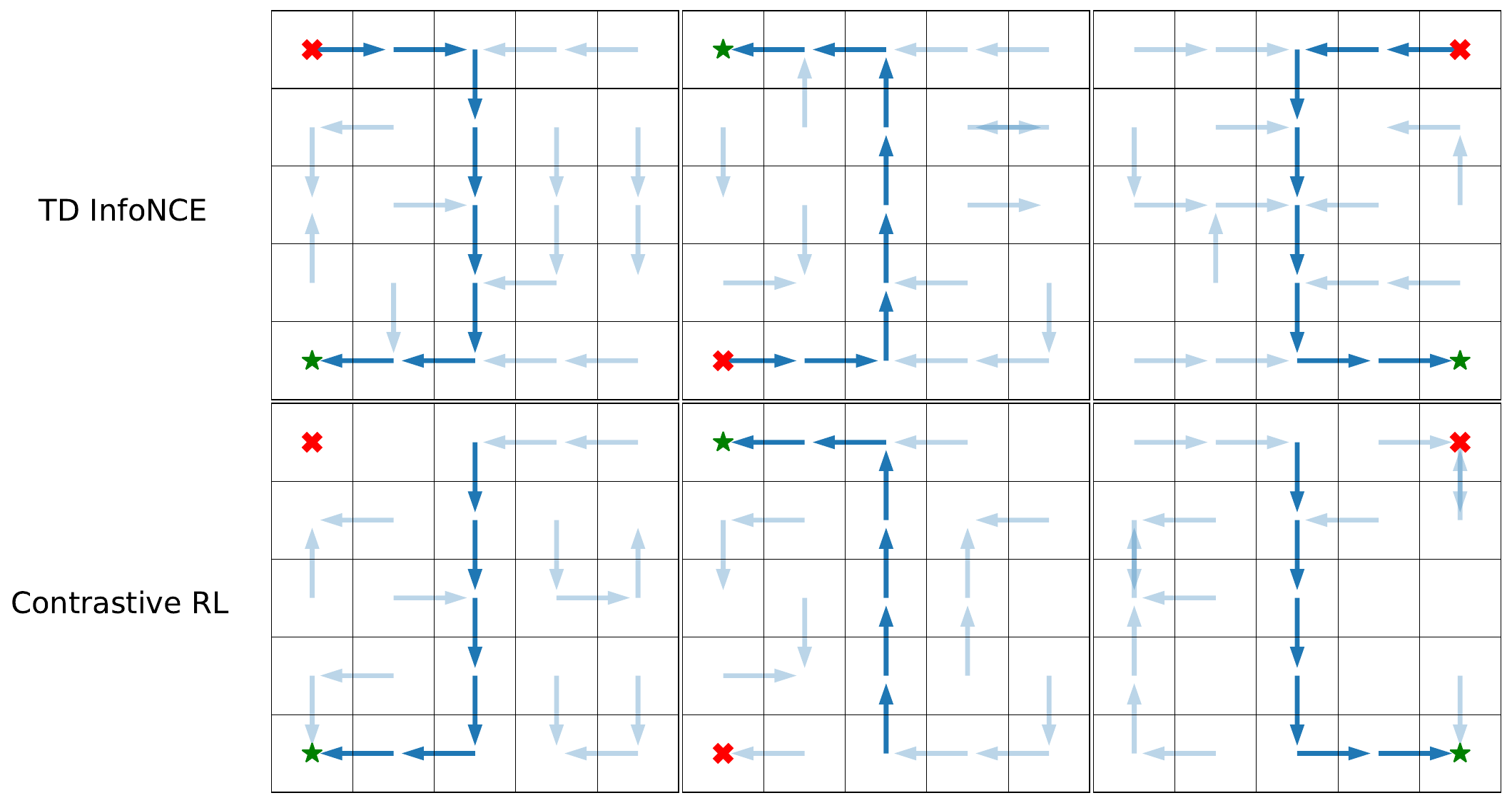}
    \end{subfigure}
    \caption{\textbf{Stitching trajectories in a dataset.} We show additional (start, goal) pairs for the experiment in Fig.~\ref{fig:stitching-property}.
    }
    \label{fig:stitching-property-more}
\end{figure*}

\paragraph{Stitching trajectories.} The first set of experiments investigate whether TD InfoNCE successfully stitches pieces of trajectories in a dataset to find complete paths between (start, goal) pairs unseen together in the dataset. We collect a dataset with size 20K consisting of "Z" style trajectories moving in diagonal and off-diagonal directions (Fig.~\ref{fig:stitching-property-more}), while evaluating the learned policy on reaching goals on the same edge as starting states after training both methods for 50K gradient steps. Figure~\ref{fig:stitching-property-more} shows that TD InfoNCE succeeds in stitching parts of trajectory in the dataset, moving along "C" style paths towards goals, while contrastive RL fails to do so. These results justify our hypothesis that TD InfoNCE performs dynamic programming and contrastive RL instead naively follows the behavior defined by the data.

\begin{figure*}
    \centering
    \includegraphics[width=\linewidth]{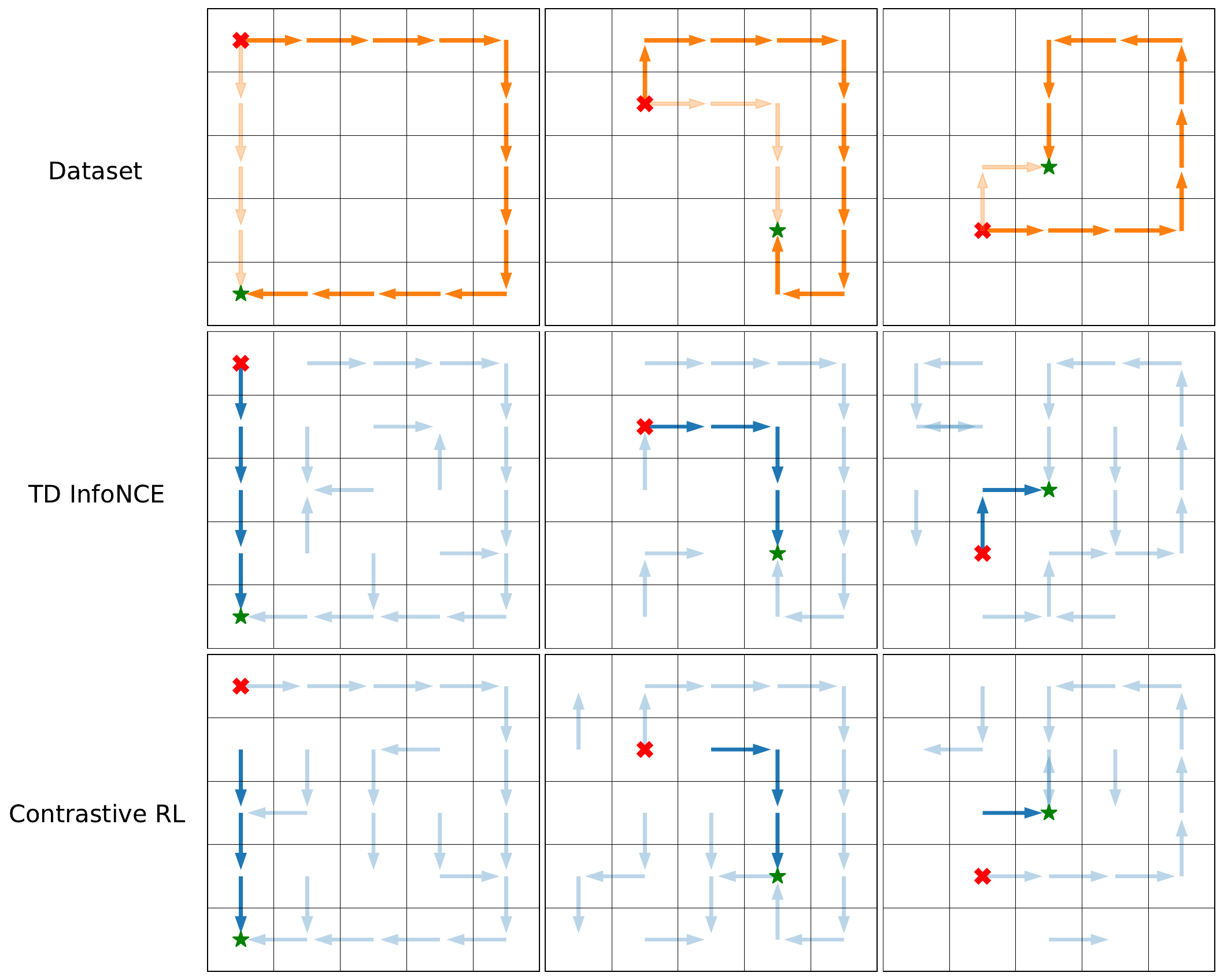}
    \caption{\textbf{Searching for shortcuts in skewed datasets.} We show additional (start, goal) pairs for the experiment in Fig.~\ref{fig:searching-shotcut}.}
    \label{fig:searching-shortcut-more}
\end{figure*}

\paragraph{Searching for shortcuts.} Our second set of experiments aim to compare the performance of TD InfoNCE against contrastive RL on searching shortcuts in skewed datasets. To study this, we collect different datasets of size 20K with trajectories conditioned on the same pair of initial state and goal, with $95\%$ of the time along a long path and $5$\% of the time along a short path. Using these skewed datasets, we again train both methods for 50K gradient steps and then evaluate the policy performance on reaching the same goal starting from the same state. We show the goal-conditioned policies learned by the two approaches in Fig.~\ref{fig:searching-shortcut-more}. The observation that TD InfoNCE learns to take shortcuts even though those data are rarely seen, while contrastive RL follows the long paths dominating the entire dataset, demonstrates the advantage of temporal difference learning over its Monte Carlo counterpart in improving data efficiency.

\subsection{Implementations and Hyperparameters}

We implement TD InfoNCE, contrastive RL, and C-Learning using JAX~\citep{bradbury2018jax} building upon the official codebase of contrastive RL\footnote{\href{https://github.com/google-research/google-research/tree/master/contrastive_rl}{https://github.com/google-research/google-research/tree/master/contrastive\_rl}}. For baselines QRL, GCBC, and DDPG + HER, we use implementation provided by the author of QRL\footnote{\href{https://github.com/quasimetric-learning/quasimetric-rl}{https://github.com/quasimetric-learning/quasimetric-rl}}. We summarize hyperparameters for TD InfoNCE in Table~\ref{tab:hparams}. Whenever possible, we used the same hyperparameters as contrastive RL~\citep{eysenbach2022contrastive}. Since TD InfoNCE computes the loss with $N^2$ negative examples, we increase the capacity of the goal-conditioned state-action encoder and the future state encoder to 4 layers MLP with 512 units in each layer applying ReLU activations. For fair comparisons, we also increased the neural network capacity in baselines to the same number and used a fixed batch size 256 for all methods. Appendix~\ref{appendix:hyperparam-ablation} includes ablations studying the effect of differet hyperparamters in Table~\ref{tab:hparams}. For offline RL experiments, we make some changes to hyperparameters (Table~\ref{tab:hparams-offline}).

\begin{table}[t]
\caption{\footnotesize Hyperparameters for TD InfoNCE.}
\label{tab:hparams}
\begin{center}
\begin{tabular}{p{8cm}|c}
\toprule
Hyperparameters & Values \\
\midrule
actor learning rate & $5 \times 10^{-5}$ \\ \midrule
critic learning rate & $3 \times 10^{-4}$ \\ \midrule
using $\ell_2$ normalized representations & yes \\ \midrule
hidden layers sizes (for both actor and representations) & $(512, 512, 512, 512)$ \\ \midrule
contrastive representation dimensions & $16$ \\
\bottomrule
\end{tabular}
\end{center}
\end{table}

\begin{table}[t]
\caption{\footnotesize Changes to hyperparameters for offline RL experiments. (Table~\ref{tab:offline-eval})}
\label{tab:hparams-offline}
\begin{center}
\begin{tabular}{p{6cm}|c}
\toprule
Hyperparameters & Values \\
\midrule
batch size (on \texttt{large-} tasks) & 256 $\to$ 1024 \\ \midrule
hidden layers sizes (for both actor and representations on \texttt{large-} tasks) & (512, 512, 512, 512) $\to$ (1024, 1024, 1024, 1024) \\ \midrule
behavioral cloning regularizer coefficient $\lambda$ & $0.1$ \\ \midrule
goals for actor loss & random states $\to$ future states \\
\bottomrule
\end{tabular}
\end{center}
\end{table}

\section{Additional Experiments}

\subsection{Hyperparameter Ablations}
\label{appendix:hyperparam-ablation}

\begin{figure*}[t]
    \centering
    \begin{subfigure}[c]{0.49\textwidth}
        \centering
        \includegraphics[width=\linewidth]{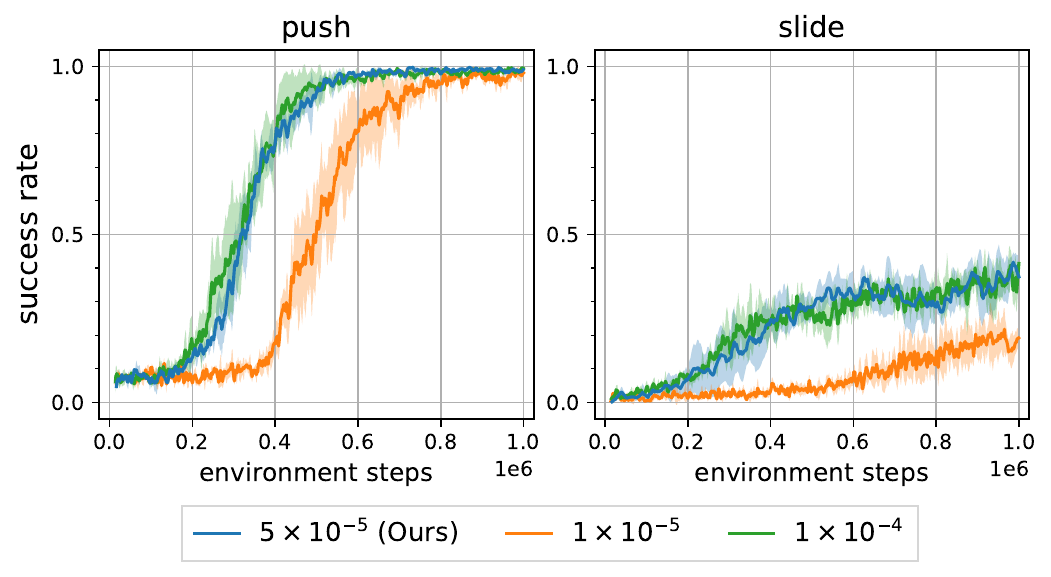}
        \caption{Actor learning rate}
    \end{subfigure}
    \hfill
    \begin{subfigure}[c]{0.49\textwidth}
        \centering
        \includegraphics[width=\linewidth]{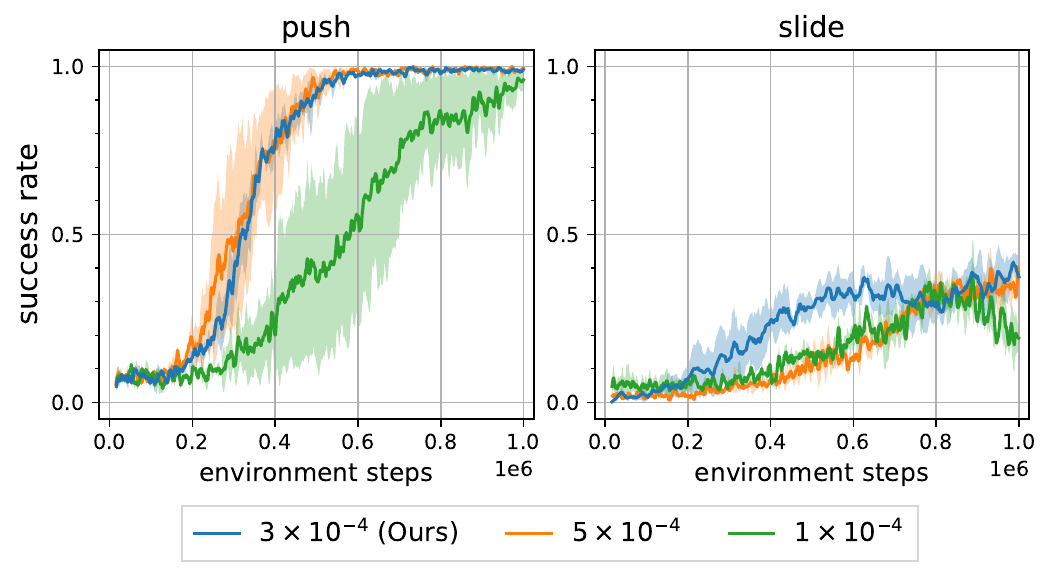}
        \caption{Critic learning rate}
    \end{subfigure}
    \vfill
    \begin{subfigure}[c]{0.49\textwidth}
        \centering
        \includegraphics[width=\linewidth]{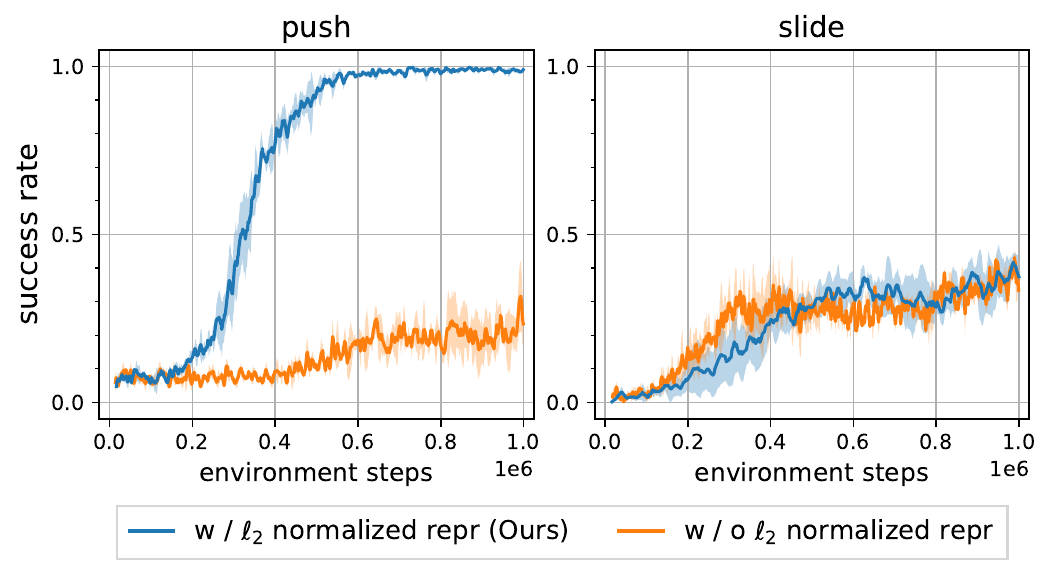}
        \caption{Representation normalization}
    \end{subfigure}
    \hfill
    \begin{subfigure}[c]{0.49\textwidth}
        \centering
        \includegraphics[width=\linewidth]{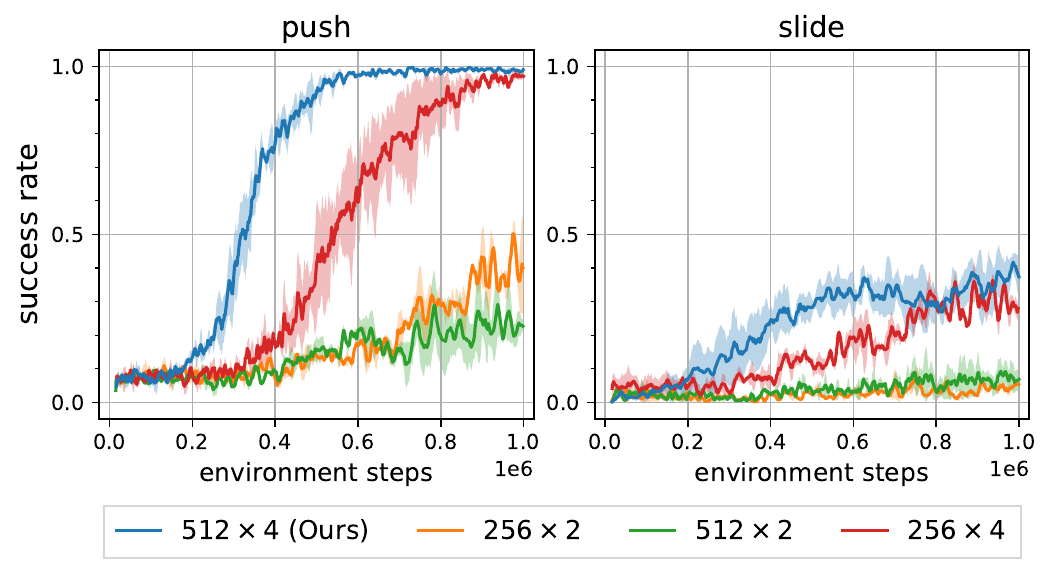}
        \caption{MLP hidden layer size\protect\footnotemark}
    \end{subfigure}
    \vfill
    \begin{subfigure}[c]{0.49\textwidth}
        \centering
        \includegraphics[width=\linewidth]{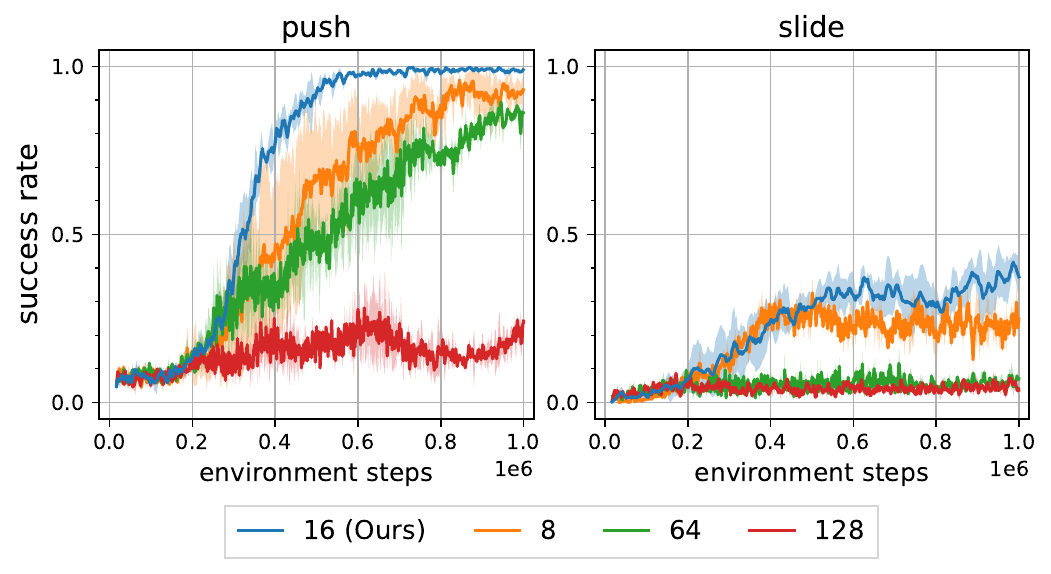}
        \caption{Representation dimension}
    \end{subfigure}
    \hfill
    \begin{subfigure}[c]{0.49\textwidth}
        \centering
        \includegraphics[width=\linewidth]{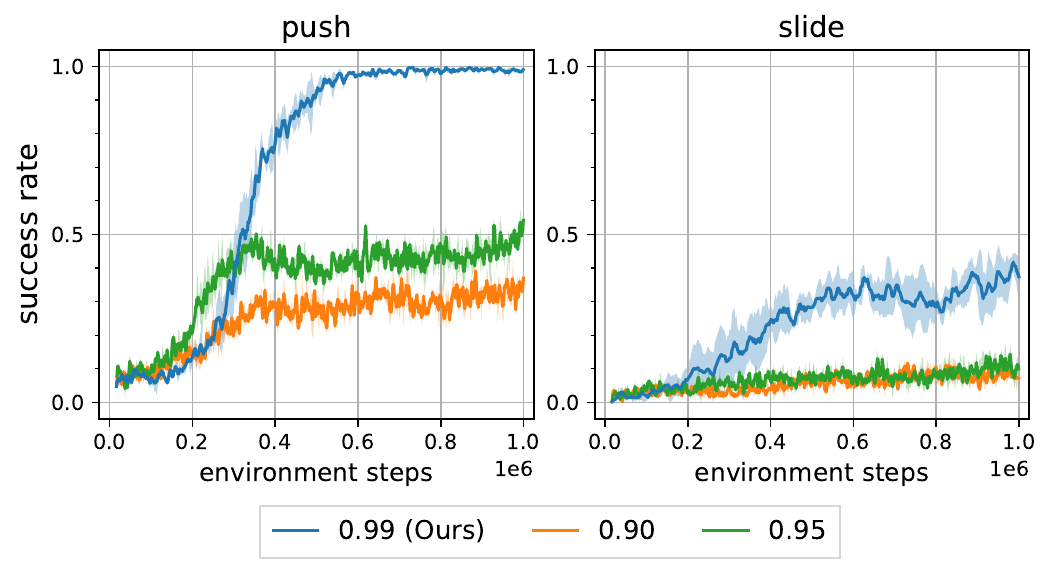}
        \caption{Discount factor $\gamma$}
    \end{subfigure}
    \caption{ \textbf{Hyperparameter ablation.} We conduct ablations to study the effect of different hyperparamters listed in Table~\ref{tab:hparams} and the discout factor $\gamma$ on state-based \texttt{push} and \texttt{slide}.
    }
    \label{fig:hyperparam-ablation}
\end{figure*}

We conduct ablation experiments to study the effect of different hyperparameters in Table~\ref{tab:hparams}, aiming to find the best hyperparameters for TD InfoNCE. For each hyperparameter, we selected a set of different values and conducted experiments on $\texttt{push (state)}$ and $\texttt{slide (state)}$, one easy task and one challenging task, for five random seeds. We report mean and standard deviation of success rates in Fig.~\ref{fig:hyperparam-ablation}. These results suggest that while some values of the hyperparameter have similar effects, e.g. actor learning rate $ = 5 \times 10^{-5}$ vs $1 \times 10^{-4}$, our choice of combination is optimal for TD InfoNCE.

\footnotetext{We use $x \times y$ to denote a $y$ layers MLP with $x$ units in each layer.}

\subsection{Predicting the discounted state occupancy measure}
\label{appendix:critic-pred-acc-full}

\begin{figure*}[t]
    \begin{minipage}{0.48\textwidth}
        \centering
        \includegraphics[width=\linewidth]{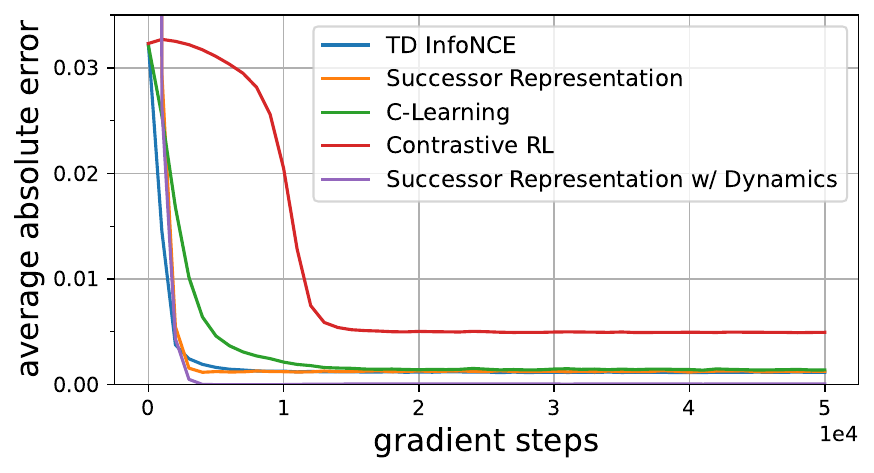}
        \vspace{-2em}
        \caption{\footnotesize Errors of discounted state occupancy measure estimation in a tabular setting.}
        \label{fig:discounted-state-occupancy-measure-est-errs-full}
    \end{minipage}
    \hfill
    \begin{minipage}{0.48\textwidth}
        \centering
        \includegraphics[width=\linewidth]{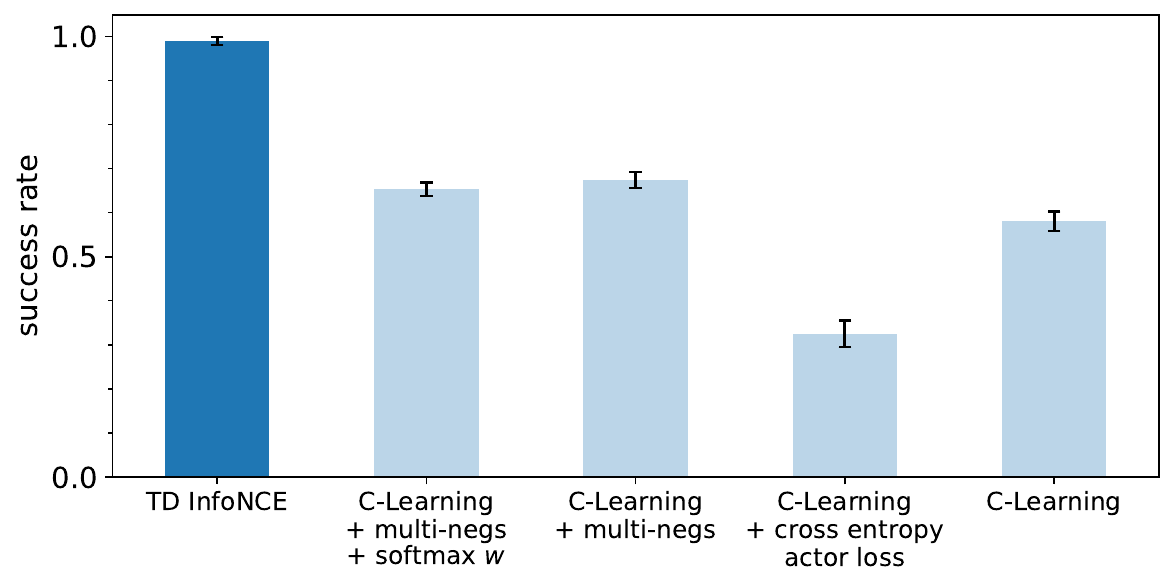}
        \caption{Differences between TD InfoNCE and C-Learning.}
        \label{fig:td-infonce-vs-c-learning}
    \end{minipage}
\end{figure*}

Our experiments estimating the discounted state occupancy measure in the tabular setting (Sec.~\ref{subsec:critic-pred-acc}) observed a small ``irreducible'' error. To test the correctness of our implementation, we applied the successor representation with a known model (Fig.~\ref{fig:discounted-state-occupancy-measure-est-errs-full}), finding that its error does go to zero. This gives us confidence that our implementation of the successor representation baseline is correct, and suggests that the error observed in Fig.~\ref{fig:discounted-state-occupancy-measure-est-errs} arises from sampling the transitions (rather than having a known model).

\subsection{Understanding the Differences between TD InfoNCE and C-Learning}
\label{appendix:td-infonce-vs-c-learning}

While conceptually similar, our method is a temporal difference estimator building upon InfoNCE whereas C-learning instead bases on the NCE objective~\citep{gutmann2010noise}. There are mainly three distinctions between TD InfoNCE and C-Learning: 
\emph{(a)} C-Learning uses a binary cross entropy loss, while TD InfoNCE uses a categorical cross entropy loss.
\emph{(b)} C-Learning uses importance weights of the form $\exp(f(s, a, g))$; if these weights are self-normalized~\citep{dubi1979interpretation, hammersley1956conditional}, they corresponds to the softmax importance weights in our objectives (Eq.~\ref{eq:importance-weight}). \emph{(c)} For the same batch of $N$ transitions, TD InfoNCE updates representations of $N^2$ negative examples (Eq.~\ref{eq:ce-negative}), while C-Learning only involves $N$ negative examples.
We ablate these decisions in Fig.~\ref{fig:td-infonce-vs-c-learning}, finding that differences (b) and (c) have little effect. Thus, we attribute the better performance of TD InfoNCE to its use of the categorical cross entropy loss.

\subsection{Representation Interpolation}
\label{appendix:latent-interp}

\begin{figure*}[t]
    \centering
    \begin{subfigure}[c]{0.49\textwidth}
    \centering
        \includegraphics[width=\linewidth]{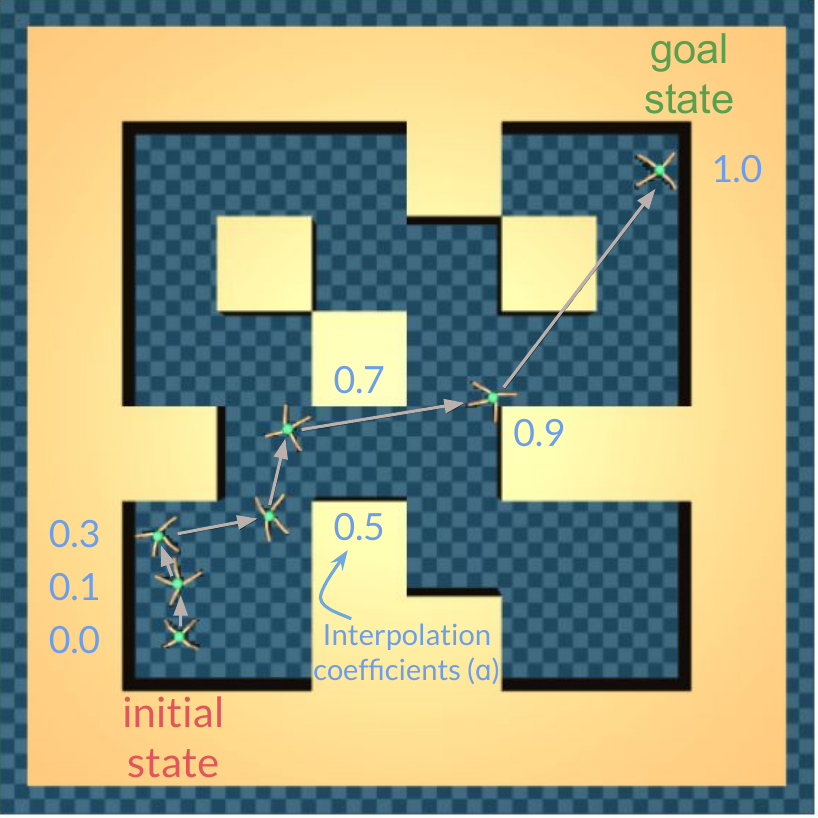}
        \caption{Parametric interpolation}
    \end{subfigure}
    \hfill
    \begin{subfigure}[c]{0.49\textwidth}
        \includegraphics[width=\linewidth]{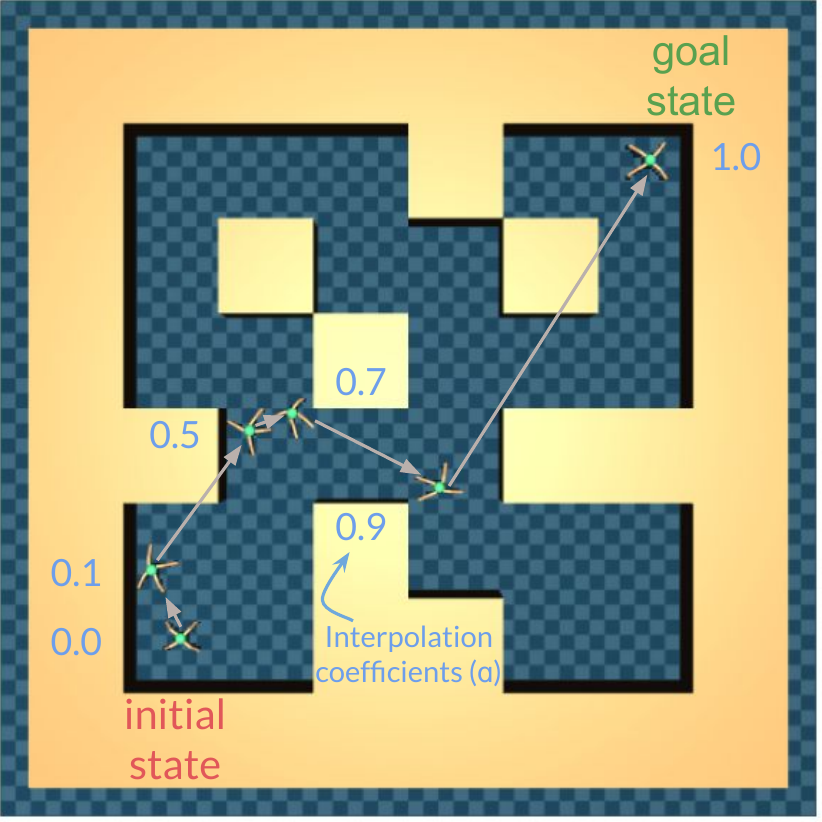}
        \caption{Non-parametric interpolation}
    \end{subfigure}
    \caption{\textbf{Visualizing representation interpolation.} Using spherical interpolation of representations \figleft~or linear interpolation of softmax features \figright, TD InfoNCE learns representations that capture not only the content of states, but also the causal relationships.
    }
    \label{fig:latent-interp}
\end{figure*}

Prior works have shown that representations learned by self-supervised learning incorporate structure of the data~\citep{wang2020understanding, arora2019theoretical}, motivating us to study whether the representations acquired by TD InfoNCE contain task-specific information. To answer this question, we visualize representations learned by TD InfoNCE via interpolating in the latent space following prior work~\citep{zheng2023stabilizing}. We choose to interpolate representations learned on the offline AntMaze \texttt{medium-play-v2} task and compare a parametric interpolation method against a non-parametric variant. Importantly, the states and goals of this task are 29 dimensions and we visualize them in 2D from a top-down view.

\paragraph{Parametric interpolation.} Given a pair of start state and goal $(s_0, g)$, we compute the normalized representations $\phi(s_0, a_{\text{no-op}}, g)$ and $\phi(g, a_{\text{no-op}}, g)$, where $a_{\text{no-op}}$ is an action taking no operation. Applying spherical linear interpolation to both of them results in blended representations, 
\begin{align*}
    \frac{\sin (1 - \alpha) \eta}{\sin \eta} \phi(s_0, a_{\text{no-op}}, g) + \frac{\sin \alpha \eta}{\sin \eta} \phi(g, a_{\text{no-op}}, g),
\end{align*}
where $\alpha \in [0, 1]$ is the interpolation coefficient and $\eta$ is the angle subtended by the arc between $\phi(s_0, a_{\text{no-op}}, g)$ and $\phi(g, a_{\text{no-op}}, g)$.
These interpolated representations can be used to find the spherical nearest neighbors in a set of representations of validation states $\{\phi(s_{\text{val}}, a_{\text{no-op}}, g) \}$ and we call this method parametric interpolation.

\paragraph{Non-parametric interpolation.} We can also sample another set of random states and using their representations $\{ \phi(s_{\text{rand}}^{(i)}, a_{\text{no-op}}, g) \}_{i = 1}^S$ as anchors to construct a softmax feature for a state $s$, $\text{feat}(s; g, \{s_{\text{rand}}\})$:
\begin{align*}
    \textsc{SoftMax} \left( \left[ \phi(s, a_{\text{no-op}}, g)^{\top} \phi(s_{\text{rand}}^{(1)}, a_{\text{no-op}}, g), \cdots, \phi(s, a_{\text{no-op}}, g)^{\top} \phi(s_{\text{rand}}^{(S)}, a_{\text{no-op}}, g) \right] \right).
\end{align*}
We compute the softmax features for representations of start and goal states and then construct the linear interpolated features,
\begin{align*}
    \alpha \text{feat}(s_0; g, \{s_{\text{rand}}\}) + (1 - \alpha) \text{feat}(g; g, \{s_{\text{rand}}\}).
\end{align*}
Those softmax features of interpolated representations are used to find the $\ell_2$ nearest neighbors in a softmax feature validation set. We call this method non-parametric interpolation. 

Results in Fig.~\ref{fig:latent-interp} suggest that when interpolating the representations using both methods, the intermediate representations correspond to sequences of states that the optimal policy should visit when reaching desired goals. Therefore, we conjecture that TD InfoNCE encodes causality in its representations while the policy learns to arrange them in a temporally correct order.


\begin{thebibliography}{97}
\providecommand{\natexlab}[1]{#1}
\providecommand{\url}[1]{\texttt{#1}}
\expandafter\ifx\csname urlstyle\endcsname\relax
  \providecommand{\doi}[1]{doi: #1}\else
  \providecommand{\doi}{doi: \begingroup \urlstyle{rm}\Url}\fi

\bibitem[Agarwal et~al.(2019)Agarwal, Jiang, Kakade, and Sun]{agarwal2019reinforcement}
Alekh Agarwal, Nan Jiang, Sham~M Kakade, and Wen Sun.
\newblock Reinforcement learning: Theory and algorithms.
\newblock \emph{CS Dept., UW Seattle, Seattle, WA, USA, Tech. Rep}, 32, 2019.

\bibitem[Andrychowicz et~al.(2017)Andrychowicz, Wolski, Ray, Schneider, Fong, Welinder, McGrew, Tobin, Pieter~Abbeel, and Zaremba]{andrychowicz2017hindsight}
Marcin Andrychowicz, Filip Wolski, Alex Ray, Jonas Schneider, Rachel Fong, Peter Welinder, Bob McGrew, Josh Tobin, OpenAI Pieter~Abbeel, and Wojciech Zaremba.
\newblock Hindsight experience replay.
\newblock \emph{Advances in neural information processing systems}, 30, 2017.

\bibitem[Arora et~al.(2019)Arora, Khandeparkar, Khodak, Plevrakis, and Saunshi]{arora2019theoretical}
Sanjeev Arora, Hrishikesh Khandeparkar, Mikhail Khodak, Orestis Plevrakis, and Nikunj Saunshi.
\newblock A theoretical analysis of contrastive unsupervised representation learning.
\newblock In \emph{36th International Conference on Machine Learning, ICML 2019}, pp.\  9904--9923. International Machine Learning Society (IMLS), 2019.

\bibitem[Barreto et~al.(2017)Barreto, Dabney, Munos, Hunt, Schaul, van Hasselt, and Silver]{barreto2017successor}
Andr{\'e} Barreto, Will Dabney, R{\'e}mi Munos, Jonathan~J Hunt, Tom Schaul, Hado~P van Hasselt, and David Silver.
\newblock Successor features for transfer in reinforcement learning.
\newblock \emph{Advances in neural information processing systems}, 30, 2017.

\bibitem[Barreto et~al.(2019)Barreto, Borsa, Hou, Comanici, Ayg{\"u}n, Hamel, Toyama, Mourad, Silver, Precup, et~al.]{barreto2019option}
Andr{\'e} Barreto, Diana Borsa, Shaobo Hou, Gheorghe Comanici, Eser Ayg{\"u}n, Philippe Hamel, Daniel Toyama, Shibl Mourad, David Silver, Doina Precup, et~al.
\newblock The option keyboard: Combining skills in reinforcement learning.
\newblock \emph{Advances in Neural Information Processing Systems}, 32, 2019.

\bibitem[Bertsekas \& Tsitsiklis(1995)Bertsekas and Tsitsiklis]{bertsekas1995neuro}
Dimitri~P Bertsekas and John~N Tsitsiklis.
\newblock Neuro-dynamic programming: an overview.
\newblock In \emph{Proceedings of 1995 34th IEEE conference on decision and control}, volume~1, pp.\  560--564. IEEE, 1995.

\bibitem[Blier et~al.(2021)Blier, Tallec, and Ollivier]{blier2021learning}
L{\'e}onard Blier, Corentin Tallec, and Yann Ollivier.
\newblock Learning successor states and goal-dependent values: A mathematical viewpoint.
\newblock \emph{arXiv preprint arXiv:2101.07123}, 2021.

\bibitem[Bradbury et~al.(2018)Bradbury, Frostig, Hawkins, Johnson, Leary, Maclaurin, Necula, Paszke, VanderPlas, Wanderman-Milne, et~al.]{bradbury2018jax}
James Bradbury, Roy Frostig, Peter Hawkins, Matthew~James Johnson, Chris Leary, Dougal Maclaurin, George Necula, Adam Paszke, Jake VanderPlas, Skye Wanderman-Milne, et~al.
\newblock Jax: composable transformations of python+ numpy programs.
\newblock 2018.

\bibitem[Campos et~al.(2020)Campos, Trott, Xiong, Socher, Gir{\'o}-i Nieto, and Torres]{campos2020explore}
V{\'\i}ctor Campos, Alexander Trott, Caiming Xiong, Richard Socher, Xavier Gir{\'o}-i Nieto, and Jordi Torres.
\newblock Explore, discover and learn: Unsupervised discovery of state-covering skills.
\newblock In \emph{International Conference on Machine Learning}, pp.\  1317--1327. PMLR, 2020.

\bibitem[Chane-Sane et~al.(2021)Chane-Sane, Schmid, and Laptev]{chane2021goal}
Elliot Chane-Sane, Cordelia Schmid, and Ivan Laptev.
\newblock Goal-conditioned reinforcement learning with imagined subgoals.
\newblock In \emph{International Conference on Machine Learning}, pp.\  1430--1440. PMLR, 2021.

\bibitem[Chen et~al.(2021)Chen, Lu, Rajeswaran, Lee, Grover, Laskin, Abbeel, Srinivas, and Mordatch]{chen2021decision}
Lili Chen, Kevin Lu, Aravind Rajeswaran, Kimin Lee, Aditya Grover, Misha Laskin, Pieter Abbeel, Aravind Srinivas, and Igor Mordatch.
\newblock Decision transformer: Reinforcement learning via sequence modeling.
\newblock \emph{Advances in neural information processing systems}, 34:\penalty0 15084--15097, 2021.

\bibitem[Chen et~al.(2020)Chen, Kornblith, Norouzi, and Hinton]{chen2020simple}
Ting Chen, Simon Kornblith, Mohammad Norouzi, and Geoffrey Hinton.
\newblock A simple framework for contrastive learning of visual representations.
\newblock In \emph{International conference on machine learning}, pp.\  1597--1607. PMLR, 2020.

\bibitem[Choi et~al.(2021)Choi, Sharma, Lee, Levine, and Gu]{choi2021variational}
Jongwook Choi, Archit Sharma, Honglak Lee, Sergey Levine, and Shixiang~Shane Gu.
\newblock Variational empowerment as representation learning for goal-conditioned reinforcement learning.
\newblock In \emph{International Conference on Machine Learning}, pp.\  1953--1963. PMLR, 2021.

\bibitem[Chopra et~al.(2005)Chopra, Hadsell, and LeCun]{chopra2005learning}
Sumit Chopra, Raia Hadsell, and Yann LeCun.
\newblock Learning a similarity metric discriminatively, with application to face verification.
\newblock In \emph{2005 IEEE computer society conference on computer vision and pattern recognition (CVPR'05)}, volume~1, pp.\  539--546. IEEE, 2005.

\bibitem[Dayan(1993)]{dayan1993improving}
Peter Dayan.
\newblock Improving generalization for temporal difference learning: The successor representation.
\newblock \emph{Neural computation}, 5\penalty0 (4):\penalty0 613--624, 1993.

\bibitem[Ding et~al.(2019)Ding, Florensa, Abbeel, and Phielipp]{ding2019goal}
Yiming Ding, Carlos Florensa, Pieter Abbeel, and Mariano Phielipp.
\newblock Goal-conditioned imitation learning.
\newblock \emph{Advances in neural information processing systems}, 32, 2019.

\bibitem[Dubi \& Horowitz(1979)Dubi and Horowitz]{dubi1979interpretation}
A~Dubi and YS~Horowitz.
\newblock The interpretation of conditional monte carlo as a form of importance sampling.
\newblock \emph{SIAM Journal on Applied Mathematics}, 36\penalty0 (1):\penalty0 115--122, 1979.

\bibitem[Durugkar et~al.(2021)Durugkar, Tec, Niekum, and Stone]{durugkar2021adversarial}
Ishan Durugkar, Mauricio Tec, Scott Niekum, and Peter Stone.
\newblock Adversarial intrinsic motivation for reinforcement learning.
\newblock \emph{Advances in Neural Information Processing Systems}, 34:\penalty0 8622--8636, 2021.

\bibitem[Emmons et~al.(2021)Emmons, Eysenbach, Kostrikov, and Levine]{emmons2021rvs}
Scott Emmons, Benjamin Eysenbach, Ilya Kostrikov, and Sergey Levine.
\newblock Rvs: What is essential for offline rl via supervised learning?
\newblock In \emph{International Conference on Learning Representations}, 2021.

\bibitem[Ernst et~al.(2005)Ernst, Geurts, and Wehenkel]{ernst2005tree}
Damien Ernst, Pierre Geurts, and Louis Wehenkel.
\newblock Tree-based batch mode reinforcement learning.
\newblock \emph{Journal of Machine Learning Research}, 6, 2005.

\bibitem[Eysenbach et~al.(2018)Eysenbach, Gupta, Ibarz, and Levine]{eysenbach2018diversity}
Benjamin Eysenbach, Abhishek Gupta, Julian Ibarz, and Sergey Levine.
\newblock Diversity is all you need: Learning skills without a reward function.
\newblock \emph{arXiv preprint arXiv:1802.06070}, 2018.

\bibitem[Eysenbach et~al.(2019)Eysenbach, Salakhutdinov, and Levine]{eysenbach2019search}
Benjamin Eysenbach, Ruslan Salakhutdinov, and Sergey Levine.
\newblock Search on the replay buffer: Bridging planning and reinforcement learning.
\newblock \emph{Advances in Neural Information Processing Systems}, 32, 2019.

\bibitem[Eysenbach et~al.(2020)Eysenbach, Salakhutdinov, and Levine]{eysenbach2020c}
Benjamin Eysenbach, Ruslan Salakhutdinov, and Sergey Levine.
\newblock C-learning: Learning to achieve goals via recursive classification.
\newblock \emph{arXiv preprint arXiv:2011.08909}, 2020.

\bibitem[Eysenbach et~al.(2022)Eysenbach, Zhang, Levine, and Salakhutdinov]{eysenbach2022contrastive}
Benjamin Eysenbach, Tianjun Zhang, Sergey Levine, and Russ~R Salakhutdinov.
\newblock Contrastive learning as goal-conditioned reinforcement learning.
\newblock \emph{Advances in Neural Information Processing Systems}, 35:\penalty0 35603--35620, 2022.

\bibitem[Fang et~al.(2022)Fang, Yin, Nair, and Levine]{fang2022planning}
Kuan Fang, Patrick Yin, Ashvin Nair, and Sergey Levine.
\newblock Planning to practice: Efficient online fine-tuning by composing goals in latent space.
\newblock In \emph{2022 IEEE/RSJ International Conference on Intelligent Robots and Systems (IROS)}, pp.\  4076--4083. IEEE, 2022.

\bibitem[Fang et~al.(2023)Fang, Yin, Nair, Walke, Yan, and Levine]{fang2023generalization}
Kuan Fang, Patrick Yin, Ashvin Nair, Homer~Rich Walke, Gengchen Yan, and Sergey Levine.
\newblock Generalization with lossy affordances: Leveraging broad offline data for learning visuomotor tasks.
\newblock In \emph{Conference on Robot Learning}, pp.\  106--117. PMLR, 2023.

\bibitem[Fu et~al.(2019)Fu, Kumar, Soh, and Levine]{fu2019diagnosing}
Justin Fu, Aviral Kumar, Matthew Soh, and Sergey Levine.
\newblock Diagnosing bottlenecks in deep q-learning algorithms.
\newblock In \emph{International Conference on Machine Learning}, pp.\  2021--2030. PMLR, 2019.

\bibitem[Fu et~al.(2020)Fu, Kumar, Nachum, Tucker, and Levine]{fu2020d4rl}
Justin Fu, Aviral Kumar, Ofir Nachum, George Tucker, and Sergey Levine.
\newblock D4rl: Datasets for deep data-driven reinforcement learning.
\newblock \emph{arXiv preprint arXiv:2004.07219}, 2020.

\bibitem[Fujimoto \& Gu(2021)Fujimoto and Gu]{fujimoto2021minimalist}
Scott Fujimoto and Shixiang~Shane Gu.
\newblock A minimalist approach to offline reinforcement learning.
\newblock \emph{Advances in neural information processing systems}, 34:\penalty0 20132--20145, 2021.

\bibitem[Gao et~al.(2021)Gao, Yao, and Chen]{gao2021simcse}
Tianyu Gao, Xingcheng Yao, and Danqi Chen.
\newblock Simcse: Simple contrastive learning of sentence embeddings.
\newblock In \emph{2021 Conference on Empirical Methods in Natural Language Processing, EMNLP 2021}, pp.\  6894--6910. Association for Computational Linguistics (ACL), 2021.

\bibitem[Gershman(2018)]{gershman2018successor}
Samuel~J Gershman.
\newblock The successor representation: its computational logic and neural substrates.
\newblock \emph{Journal of Neuroscience}, 38\penalty0 (33):\penalty0 7193--7200, 2018.

\bibitem[Ghosh et~al.(2020)Ghosh, Gupta, Reddy, Fu, Devin, Eysenbach, and Levine]{ghosh2020learning}
Dibya Ghosh, Abhishek Gupta, Ashwin Reddy, Justin Fu, Coline~Manon Devin, Benjamin Eysenbach, and Sergey Levine.
\newblock Learning to reach goals via iterated supervised learning.
\newblock In \emph{International Conference on Learning Representations}, 2020.

\bibitem[Giles(2015)]{giles2015multilevel}
Michael~B Giles.
\newblock Multilevel monte carlo methods.
\newblock \emph{Acta numerica}, 24:\penalty0 259--328, 2015.

\bibitem[Gregor et~al.(2016)Gregor, Rezende, and Wierstra]{gregor2016variational}
Karol Gregor, Danilo~Jimenez Rezende, and Daan Wierstra.
\newblock Variational intrinsic control.
\newblock \emph{arXiv preprint arXiv:1611.07507}, 2016.

\bibitem[Grill et~al.(2020)Grill, Strub, Altch{\'e}, Tallec, Richemond, Buchatskaya, Doersch, Avila~Pires, Guo, Gheshlaghi~Azar, et~al.]{grill2020bootstrap}
Jean-Bastien Grill, Florian Strub, Florent Altch{\'e}, Corentin Tallec, Pierre Richemond, Elena Buchatskaya, Carl Doersch, Bernardo Avila~Pires, Zhaohan Guo, Mohammad Gheshlaghi~Azar, et~al.
\newblock Bootstrap your own latent-a new approach to self-supervised learning.
\newblock \emph{Advances in neural information processing systems}, 33:\penalty0 21271--21284, 2020.

\bibitem[Gupta et~al.(2020)Gupta, Kumar, Lynch, Levine, and Hausman]{gupta2020relay}
Abhishek Gupta, Vikash Kumar, Corey Lynch, Sergey Levine, and Karol Hausman.
\newblock Relay policy learning: Solving long-horizon tasks via imitation and reinforcement learning.
\newblock In \emph{Conference on Robot Learning}, pp.\  1025--1037. PMLR, 2020.

\bibitem[Gutmann \& Hyv{\"a}rinen(2010)Gutmann and Hyv{\"a}rinen]{gutmann2010noise}
Michael Gutmann and Aapo Hyv{\"a}rinen.
\newblock Noise-contrastive estimation: A new estimation principle for unnormalized statistical models.
\newblock In \emph{Proceedings of the thirteenth international conference on artificial intelligence and statistics}, pp.\  297--304. JMLR Workshop and Conference Proceedings, 2010.

\bibitem[Hammersley(1956)]{hammersley1956conditional}
JM~Hammersley.
\newblock Conditional monte carlo.
\newblock \emph{Journal of the ACM (JACM)}, 3\penalty0 (2):\penalty0 73--76, 1956.

\bibitem[Hansen-Estruch et~al.(2022)Hansen-Estruch, Zhang, Nair, Yin, and Levine]{hansen2022bisimulation}
Philippe Hansen-Estruch, Amy Zhang, Ashvin Nair, Patrick Yin, and Sergey Levine.
\newblock Bisimulation makes analogies in goal-conditioned reinforcement learning.
\newblock In \emph{International Conference on Machine Learning}, pp.\  8407--8426. PMLR, 2022.

\bibitem[He et~al.(2022)He, Chen, Xie, Li, Doll{\'a}r, and Girshick]{he2022masked}
Kaiming He, Xinlei Chen, Saining Xie, Yanghao Li, Piotr Doll{\'a}r, and Ross Girshick.
\newblock Masked autoencoders are scalable vision learners.
\newblock In \emph{Proceedings of the IEEE/CVF conference on computer vision and pattern recognition}, pp.\  16000--16009, 2022.

\bibitem[Henaff(2020)]{henaff2020data}
Olivier Henaff.
\newblock Data-efficient image recognition with contrastive predictive coding.
\newblock In \emph{International conference on machine learning}, pp.\  4182--4192. PMLR, 2020.

\bibitem[Ho \& Ermon(2016)Ho and Ermon]{ho2016generative}
Jonathan Ho and Stefano Ermon.
\newblock Generative adversarial imitation learning.
\newblock \emph{Advances in neural information processing systems}, 29, 2016.

\bibitem[Janner et~al.(2020)Janner, Mordatch, and Levine]{janner2020gamma}
Michael Janner, Igor Mordatch, and Sergey Levine.
\newblock gamma-models: Generative temporal difference learning for infinite-horizon prediction.
\newblock \emph{Advances in Neural Information Processing Systems}, 33:\penalty0 1724--1735, 2020.

\bibitem[Jia et~al.(2021)Jia, Yang, Xia, Chen, Parekh, Pham, Le, Sung, Li, and Duerig]{jia2021scaling}
Chao Jia, Yinfei Yang, Ye~Xia, Yi-Ting Chen, Zarana Parekh, Hieu Pham, Quoc Le, Yun-Hsuan Sung, Zhen Li, and Tom Duerig.
\newblock Scaling up visual and vision-language representation learning with noisy text supervision.
\newblock In \emph{International conference on machine learning}, pp.\  4904--4916. PMLR, 2021.

\bibitem[Jozefowicz et~al.(2016)Jozefowicz, Vinyals, Schuster, Shazeer, and Wu]{jozefowicz2016exploring}
Rafal Jozefowicz, Oriol Vinyals, Mike Schuster, Noam Shazeer, and Yonghui Wu.
\newblock Exploring the limits of language modeling.
\newblock \emph{arXiv preprint arXiv:1602.02410}, 2016.

\bibitem[Kostrikov et~al.(2021)Kostrikov, Nair, and Levine]{kostrikov2021offline}
Ilya Kostrikov, Ashvin Nair, and Sergey Levine.
\newblock Offline reinforcement learning with implicit q-learning.
\newblock In \emph{International Conference on Learning Representations}, 2021.

\bibitem[Kumar et~al.(2019)Kumar, Fu, Soh, Tucker, and Levine]{kumar2019stabilizing}
Aviral Kumar, Justin Fu, Matthew Soh, George Tucker, and Sergey Levine.
\newblock Stabilizing off-policy q-learning via bootstrapping error reduction.
\newblock \emph{Advances in Neural Information Processing Systems}, 32, 2019.

\bibitem[Kumar et~al.(2020)Kumar, Zhou, Tucker, and Levine]{kumar2020conservative}
Aviral Kumar, Aurick Zhou, George Tucker, and Sergey Levine.
\newblock Conservative q-learning for offline reinforcement learning.
\newblock \emph{Advances in Neural Information Processing Systems}, 33:\penalty0 1179--1191, 2020.

\bibitem[Laskin et~al.(2020{\natexlab{a}})Laskin, Srinivas, and Abbeel]{laskin2020curl}
Michael Laskin, Aravind Srinivas, and Pieter Abbeel.
\newblock Curl: Contrastive unsupervised representations for reinforcement learning.
\newblock In \emph{International Conference on Machine Learning}, pp.\  5639--5650. PMLR, 2020{\natexlab{a}}.

\bibitem[Laskin et~al.(2020{\natexlab{b}})Laskin, Lee, Stooke, Pinto, Abbeel, and Srinivas]{laskin2020reinforcement}
Misha Laskin, Kimin Lee, Adam Stooke, Lerrel Pinto, Pieter Abbeel, and Aravind Srinivas.
\newblock Reinforcement learning with augmented data.
\newblock \emph{Advances in neural information processing systems}, 33:\penalty0 19884--19895, 2020{\natexlab{b}}.

\bibitem[Levy et~al.(2018)Levy, Konidaris, Platt, and Saenko]{levy2018learning}
Andrew Levy, George Konidaris, Robert Platt, and Kate Saenko.
\newblock Learning multi-level hierarchies with hindsight.
\newblock In \emph{International Conference on Learning Representations}, 2018.

\bibitem[Linsker(1988)]{linsker1988self}
Ralph Linsker.
\newblock Self-organization in a perceptual network.
\newblock \emph{Computer}, 21\penalty0 (3):\penalty0 105--117, 1988.

\bibitem[Logeswaran \& Lee(2018)Logeswaran and Lee]{logeswaran2018efficient}
Lajanugen Logeswaran and Honglak Lee.
\newblock An efficient framework for learning sentence representations.
\newblock \emph{arXiv preprint arXiv:1803.02893}, 2018.

\bibitem[Lynch et~al.(2020)Lynch, Khansari, Xiao, Kumar, Tompson, Levine, and Sermanet]{lynch2020learning}
Corey Lynch, Mohi Khansari, Ted Xiao, Vikash Kumar, Jonathan Tompson, Sergey Levine, and Pierre Sermanet.
\newblock Learning latent plans from play.
\newblock In \emph{Conference on robot learning}, pp.\  1113--1132. PMLR, 2020.

\bibitem[Ma et~al.(2022)Ma, Sodhani, Jayaraman, Bastani, Kumar, and Zhang]{ma2022vip}
Yecheng~Jason Ma, Shagun Sodhani, Dinesh Jayaraman, Osbert Bastani, Vikash Kumar, and Amy Zhang.
\newblock Vip: Towards universal visual reward and representation via value-implicit pre-training.
\newblock In \emph{The Eleventh International Conference on Learning Representations}, 2022.

\bibitem[Ma \& Collins(2018)Ma and Collins]{ma2018noise}
Zhuang Ma and Michael Collins.
\newblock Noise contrastive estimation and negative sampling for conditional models: Consistency and statistical efficiency.
\newblock \emph{arXiv preprint arXiv:1809.01812}, 2018.

\bibitem[Mazoure et~al.(2022)Mazoure, Eysenbach, Nachum, and Tompson]{mazoure2022contrastive}
Bogdan Mazoure, Benjamin Eysenbach, Ofir Nachum, and Jonathan Tompson.
\newblock Contrastive value learning: Implicit models for simple offline rl.
\newblock \emph{arXiv preprint arXiv:2211.02100}, 2022.

\bibitem[Mnih et~al.(2015)Mnih, Kavukcuoglu, Silver, Rusu, Veness, Bellemare, Graves, Riedmiller, Fidjeland, Ostrovski, et~al.]{mnih2015human}
Volodymyr Mnih, Koray Kavukcuoglu, David Silver, Andrei~A Rusu, Joel Veness, Marc~G Bellemare, Alex Graves, Martin Riedmiller, Andreas~K Fidjeland, Georg Ostrovski, et~al.
\newblock Human-level control through deep reinforcement learning.
\newblock \emph{nature}, 518\penalty0 (7540):\penalty0 529--533, 2015.

\bibitem[Nachum et~al.(2018)Nachum, Gu, Lee, and Levine]{nachum2018data}
Ofir Nachum, Shixiang~Shane Gu, Honglak Lee, and Sergey Levine.
\newblock Data-efficient hierarchical reinforcement learning.
\newblock \emph{Advances in neural information processing systems}, 31, 2018.

\bibitem[Nair et~al.(2020{\natexlab{a}})Nair, Bahl, Khazatsky, Pong, Berseth, and Levine]{nair2020contextual}
Ashvin Nair, Shikhar Bahl, Alexander Khazatsky, Vitchyr Pong, Glen Berseth, and Sergey Levine.
\newblock Contextual imagined goals for self-supervised robotic learning.
\newblock In \emph{Conference on Robot Learning}, pp.\  530--539. PMLR, 2020{\natexlab{a}}.

\bibitem[Nair et~al.(2018)Nair, Pong, Dalal, Bahl, Lin, and Levine]{nair2018visual}
Ashvin~V Nair, Vitchyr Pong, Murtaza Dalal, Shikhar Bahl, Steven Lin, and Sergey Levine.
\newblock Visual reinforcement learning with imagined goals.
\newblock \emph{Advances in neural information processing systems}, 31, 2018.

\bibitem[Nair \& Finn(2019)Nair and Finn]{nair2019hierarchical}
Suraj Nair and Chelsea Finn.
\newblock Hierarchical foresight: Self-supervised learning of long-horizon tasks via visual subgoal generation.
\newblock In \emph{International Conference on Learning Representations}, 2019.

\bibitem[Nair et~al.(2020{\natexlab{b}})Nair, Savarese, and Finn]{nair2020goal}
Suraj Nair, Silvio Savarese, and Chelsea Finn.
\newblock Goal-aware prediction: Learning to model what matters.
\newblock In \emph{International Conference on Machine Learning}, pp.\  7207--7219. PMLR, 2020{\natexlab{b}}.

\bibitem[Nair et~al.(2022)Nair, Rajeswaran, Kumar, Finn, and Gupta]{nair2022r3m}
Suraj Nair, Aravind Rajeswaran, Vikash Kumar, Chelsea Finn, and Abhinav Gupta.
\newblock R3m: A universal visual representation for robot manipulation.
\newblock \emph{arXiv preprint arXiv:2203.12601}, 2022.

\bibitem[Oh et~al.(2018)Oh, Guo, Singh, and Lee]{oh2018self}
Junhyuk Oh, Yijie Guo, Satinder Singh, and Honglak Lee.
\newblock Self-imitation learning.
\newblock In \emph{International Conference on Machine Learning}, pp.\  3878--3887. PMLR, 2018.

\bibitem[Oh~Song et~al.(2016)Oh~Song, Xiang, Jegelka, and Savarese]{oh2016deep}
Hyun Oh~Song, Yu~Xiang, Stefanie Jegelka, and Silvio Savarese.
\newblock Deep metric learning via lifted structured feature embedding.
\newblock In \emph{Proceedings of the IEEE conference on computer vision and pattern recognition}, pp.\  4004--4012, 2016.

\bibitem[Oord et~al.(2018)Oord, Li, and Vinyals]{oord2018representation}
Aaron van~den Oord, Yazhe Li, and Oriol Vinyals.
\newblock Representation learning with contrastive predictive coding.
\newblock \emph{arXiv preprint arXiv:1807.03748}, 2018.

\bibitem[Pertsch et~al.(2020)Pertsch, Rybkin, Ebert, Zhou, Jayaraman, Finn, and Levine]{pertsch2020long}
Karl Pertsch, Oleh Rybkin, Frederik Ebert, Shenghao Zhou, Dinesh Jayaraman, Chelsea Finn, and Sergey Levine.
\newblock Long-horizon visual planning with goal-conditioned hierarchical predictors.
\newblock \emph{Advances in Neural Information Processing Systems}, 33:\penalty0 17321--17333, 2020.

\bibitem[Plappert et~al.(2018)Plappert, Andrychowicz, Ray, McGrew, Baker, Powell, Schneider, Tobin, Chociej, Welinder, et~al.]{plappert2018multi}
Matthias Plappert, Marcin Andrychowicz, Alex Ray, Bob McGrew, Bowen Baker, Glenn Powell, Jonas Schneider, Josh Tobin, Maciek Chociej, Peter Welinder, et~al.
\newblock Multi-goal reinforcement learning: Challenging robotics environments and request for research.
\newblock \emph{arXiv preprint arXiv:1802.09464}, 2018.

\bibitem[Poole et~al.(2019)Poole, Ozair, Van Den~Oord, Alemi, and Tucker]{poole2019variational}
Ben Poole, Sherjil Ozair, Aaron Van Den~Oord, Alex Alemi, and George Tucker.
\newblock On variational bounds of mutual information.
\newblock In \emph{International Conference on Machine Learning}, pp.\  5171--5180. PMLR, 2019.

\bibitem[Precup et~al.(2000)Precup, Sutton, and Singh]{precup2000eligibility}
Doina Precup, Richard~S Sutton, and Satinder~P Singh.
\newblock Eligibility traces for off-policy policy evaluation.
\newblock In \emph{Proceedings of the Seventeenth International Conference on Machine Learning}, pp.\  759--766, 2000.

\bibitem[Precup et~al.(2001)Precup, Sutton, and Dasgupta]{precup2001off}
Doina Precup, Richard~S Sutton, and Sanjoy Dasgupta.
\newblock Off-policy temporal difference learning with function approximation.
\newblock In \emph{Proceedings of the Eighteenth International Conference on Machine Learning}, pp.\  417--424, 2001.

\bibitem[Radford et~al.(2021)Radford, Kim, Hallacy, Ramesh, Goh, Agarwal, Sastry, Askell, Mishkin, Clark, et~al.]{radford2021learning}
Alec Radford, Jong~Wook Kim, Chris Hallacy, Aditya Ramesh, Gabriel Goh, Sandhini Agarwal, Girish Sastry, Amanda Askell, Pamela Mishkin, Jack Clark, et~al.
\newblock Learning transferable visual models from natural language supervision.
\newblock In \emph{International conference on machine learning}, pp.\  8748--8763. PMLR, 2021.

\bibitem[Rainforth et~al.(2018)Rainforth, Cornish, Yang, Warrington, and Wood]{rainforth2018nesting}
Tom Rainforth, Rob Cornish, Hongseok Yang, Andrew Warrington, and Frank Wood.
\newblock On nesting monte carlo estimators.
\newblock In \emph{International Conference on Machine Learning}, pp.\  4267--4276. PMLR, 2018.

\bibitem[Riedmiller et~al.(2018)Riedmiller, Hafner, Lampe, Neunert, Degrave, Wiele, Mnih, Heess, and Springenberg]{riedmiller2018learning}
Martin Riedmiller, Roland Hafner, Thomas Lampe, Michael Neunert, Jonas Degrave, Tom Wiele, Vlad Mnih, Nicolas Heess, and Jost~Tobias Springenberg.
\newblock Learning by playing solving sparse reward tasks from scratch.
\newblock In \emph{International conference on machine learning}, pp.\  4344--4353. PMLR, 2018.

\bibitem[Rudner et~al.(2021)Rudner, Pong, McAllister, Gal, and Levine]{rudner2021outcome}
Tim~GJ Rudner, Vitchyr Pong, Rowan McAllister, Yarin Gal, and Sergey Levine.
\newblock Outcome-driven reinforcement learning via variational inference.
\newblock \emph{Advances in Neural Information Processing Systems}, 34:\penalty0 13045--13058, 2021.

\bibitem[Schaul et~al.(2015)Schaul, Horgan, Gregor, and Silver]{schaul2015universal}
Tom Schaul, Daniel Horgan, Karol Gregor, and David Silver.
\newblock Universal value function approximators.
\newblock In \emph{International conference on machine learning}, pp.\  1312--1320. PMLR, 2015.

\bibitem[Schroff et~al.(2015)Schroff, Kalenichenko, and Philbin]{schroff2015facenet}
Florian Schroff, Dmitry Kalenichenko, and James Philbin.
\newblock Facenet: A unified embedding for face recognition and clustering.
\newblock In \emph{Proceedings of the IEEE conference on computer vision and pattern recognition}, pp.\  815--823, 2015.

\bibitem[Sermanet et~al.(2018)Sermanet, Lynch, Chebotar, Hsu, Jang, Schaal, Levine, and Brain]{sermanet2018time}
Pierre Sermanet, Corey Lynch, Yevgen Chebotar, Jasmine Hsu, Eric Jang, Stefan Schaal, Sergey Levine, and Google Brain.
\newblock Time-contrastive networks: Self-supervised learning from video.
\newblock In \emph{2018 IEEE international conference on robotics and automation (ICRA)}, pp.\  1134--1141. IEEE, 2018.

\bibitem[Shah et~al.(2022)Shah, Eysenbach, Rhinehart, and Levine]{shah2022rapid}
Dhruv Shah, Benjamin Eysenbach, Nicholas Rhinehart, and Sergey Levine.
\newblock Rapid exploration for open-world navigation with latent goal models.
\newblock In \emph{Conference on Robot Learning}, pp.\  674--684. PMLR, 2022.

\bibitem[Sohn(2016)]{sohn2016improved}
Kihyuk Sohn.
\newblock Improved deep metric learning with multi-class n-pair loss objective.
\newblock \emph{Advances in neural information processing systems}, 29, 2016.

\bibitem[Srivastava et~al.(2019)Srivastava, Shyam, Mutz, Ja{\'s}kowski, and Schmidhuber]{srivastava2019training}
Rupesh~Kumar Srivastava, Pranav Shyam, Filipe Mutz, Wojciech Ja{\'s}kowski, and J{\"u}rgen Schmidhuber.
\newblock Training agents using upside-down reinforcement learning.
\newblock \emph{arXiv preprint arXiv:1912.02877}, 2019.

\bibitem[Sun et~al.(2019)Sun, Li, Liu, Zhou, and Lin]{sun2019policy}
Hao Sun, Zhizhong Li, Xiaotong Liu, Bolei Zhou, and Dahua Lin.
\newblock Policy continuation with hindsight inverse dynamics.
\newblock \emph{Advances in Neural Information Processing Systems}, 32, 2019.

\bibitem[Sutton \& Barto(2018)Sutton and Barto]{sutton2018reinforcement}
Richard~S Sutton and Andrew~G Barto.
\newblock \emph{Reinforcement learning: An introduction}.
\newblock MIT press, 2018.

\bibitem[Tian et~al.(2020{\natexlab{a}})Tian, Nair, Ebert, Dasari, Eysenbach, Finn, and Levine]{tian2020model}
Stephen Tian, Suraj Nair, Frederik Ebert, Sudeep Dasari, Benjamin Eysenbach, Chelsea Finn, and Sergey Levine.
\newblock Model-based visual planning with self-supervised functional distances.
\newblock In \emph{International Conference on Learning Representations}, 2020{\natexlab{a}}.

\bibitem[Tian et~al.(2020{\natexlab{b}})Tian, Krishnan, and Isola]{tian2020contrastive}
Yonglong Tian, Dilip Krishnan, and Phillip Isola.
\newblock Contrastive multiview coding.
\newblock In \emph{Computer Vision--ECCV 2020: 16th European Conference, Glasgow, UK, August 23--28, 2020, Proceedings, Part XI 16}, pp.\  776--794. Springer, 2020{\natexlab{b}}.

\bibitem[Touati \& Ollivier(2021)Touati and Ollivier]{touati2021learning}
Ahmed Touati and Yann Ollivier.
\newblock Learning one representation to optimize all rewards.
\newblock \emph{Advances in Neural Information Processing Systems}, 34:\penalty0 13--23, 2021.

\bibitem[Tsai et~al.(2020)Tsai, Zhao, Yamada, Morency, and Salakhutdinov]{tsai2020neural}
Yao-Hung~Hubert Tsai, Han Zhao, Makoto Yamada, Louis-Philippe Morency, and Russ~R Salakhutdinov.
\newblock Neural methods for point-wise dependency estimation.
\newblock \emph{Advances in Neural Information Processing Systems}, 33:\penalty0 62--72, 2020.

\bibitem[Tschannen et~al.(2019)Tschannen, Djolonga, Rubenstein, Gelly, and Lucic]{tschannen2019mutual}
Michael Tschannen, Josip Djolonga, Paul~K Rubenstein, Sylvain Gelly, and Mario Lucic.
\newblock On mutual information maximization for representation learning.
\newblock \emph{arXiv preprint arXiv:1907.13625}, 2019.

\bibitem[Wang \& Isola(2020)Wang and Isola]{wang2020understanding}
Tongzhou Wang and Phillip Isola.
\newblock Understanding contrastive representation learning through alignment and uniformity on the hypersphere.
\newblock In \emph{International Conference on Machine Learning}, pp.\  9929--9939. PMLR, 2020.

\bibitem[Wang et~al.(2023)Wang, Torralba, Isola, and Zhang]{pmlr-v202-wang23al}
Tongzhou Wang, Antonio Torralba, Phillip Isola, and Amy Zhang.
\newblock Optimal goal-reaching reinforcement learning via quasimetric learning.
\newblock In Andreas Krause, Emma Brunskill, Kyunghyun Cho, Barbara Engelhardt, Sivan Sabato, and Jonathan Scarlett (eds.), \emph{Proceedings of the 40th International Conference on Machine Learning}, volume 202 of \emph{Proceedings of Machine Learning Research}, pp.\  36411--36430. PMLR, 23--29 Jul 2023.
\newblock URL \url{https://proceedings.mlr.press/v202/wang23al.html}.

\bibitem[Warde-Farley et~al.(2018)Warde-Farley, Van~de Wiele, Kulkarni, Ionescu, Hansen, and Mnih]{warde2018unsupervised}
David Warde-Farley, Tom Van~de Wiele, Tejas Kulkarni, Catalin Ionescu, Steven Hansen, and Volodymyr Mnih.
\newblock Unsupervised control through non-parametric discriminative rewards.
\newblock \emph{arXiv preprint arXiv:1811.11359}, 2018.

\bibitem[Watkins \& Dayan(1992)Watkins and Dayan]{watkins1992q}
Christopher~JCH Watkins and Peter Dayan.
\newblock Q-learning.
\newblock \emph{Machine learning}, 8:\penalty0 279--292, 1992.

\bibitem[Weinberger \& Saul(2009)Weinberger and Saul]{weinberger2009distance}
Kilian~Q Weinberger and Lawrence~K Saul.
\newblock Distance metric learning for large margin nearest neighbor classification.
\newblock \emph{Journal of machine learning research}, 10\penalty0 (2), 2009.

\bibitem[Wu et~al.(2018)Wu, Xiong, Yu, and Lin]{wu2018unsupervised}
Zhirong Wu, Yuanjun Xiong, Stella~X Yu, and Dahua Lin.
\newblock Unsupervised feature learning via non-parametric instance discrimination.
\newblock In \emph{Proceedings of the IEEE conference on computer vision and pattern recognition}, pp.\  3733--3742, 2018.

\bibitem[Zhang et~al.(2020)Zhang, Liu, and Whiteson]{zhang2020gradientdice}
Shangtong Zhang, Bo~Liu, and Shimon Whiteson.
\newblock Gradientdice: Rethinking generalized offline estimation of stationary values.
\newblock In \emph{International Conference on Machine Learning}, pp.\  11194--11203. PMLR, 2020.

\bibitem[Zheng et~al.(2023)Zheng, Eysenbach, Walke, Yin, Fang, Salakhutdinov, and Levine]{zheng2023stabilizing}
Chongyi Zheng, Benjamin Eysenbach, Homer Walke, Patrick Yin, Kuan Fang, Ruslan Salakhutdinov, and Sergey Levine.
\newblock Stabilizing contrastive rl: Techniques for offline goal reaching.
\newblock \emph{arXiv preprint arXiv:2306.03346}, 2023.

\end{thebibliography}
\end{document}